\documentclass[runningheads]{llncs}

\usepackage{times}
\usepackage{epsfig}
\usepackage{graphicx}
\usepackage{amsmath}
\usepackage{amssymb}
\usepackage{xspace}
\usepackage{lineno}
\usepackage[pagebackref=true,breaklinks=true,colorlinks,bookmarks=false]{hyperref}
\usepackage[normalem]{ulem}
\usepackage[toc,page]{appendix}
\usepackage{wrapfig}

\usepackage[numbers,sort,compress]{natbib}

\newcommand{\ourNUmbColor}{\textcolor{black}}

\newcommand{\ourNumbCam}{\ourNUmbColor{$6$}\xspace}
\newcommand{\ourNumbObj}{{\ourNUmbColor{10}}\xspace}
\newcommand{\ourNumbSubj}{\ourNUmbColor{10}\xspace}
\newcommand{\ourNumbSubjM}{\ourNUmbColor{5}\xspace}
\newcommand{\ourNumbSubjF}{\ourNUmbColor{5}\xspace}
\newcommand{\ourNumbMotions}{{\ourNUmbColor{223}}\xspace}
\newcommand{\ourNumbFrames}{{\ourNUmbColor{67,357}}\xspace}

\newcommand{\supmat}{\textcolor{black}{{Appx.}}\xspace}

\newcommand{\kinectAZ}{{Azure Kinect}\xspace}
\newcommand{\kinectAZs}{{Azure Kinects}\xspace}
\newcommand{\pointrend}{{PointRend}\xspace}

\newcommand{\imus}{{IMUs}\xspace}

\newcommand{\REBT}[1]{\xspace{\color{black} #1}\xspace}
\newcommand{\GCPR}[1]{\xspace{\color{black} #1}\xspace}

\usepackage{pifont}
\newcommand{\cmark}{\color{green}\ding{51}}
\newcommand{\xmark}{\color{red}\ding{55}}

\newcommand{\mAcc}{\mathcal{S}}
\newcommand{\mAccc}{\mathcal{A}}

\newcommand{\websiteURL}{\mbox{\url{https://intercap.is.tue.mpg.de}}}

\newcommand{\ourrefcolor}{\textcolor{black}}

\newcommand{\tab}[1]{\mbox{\ourrefcolor{{Tab.}}~\ref{#1}}}
\newcommand{\fig}[1]{\mbox{\ourrefcolor{{Fig.}}~\ref{#1}}}

\newcommand{\Fig}[1]{\mbox{\ourrefcolor{{Figure}}~\ref{#1}}}

\newcommand{\intercap}{InterCap\xspace}
\newcommand{\ourname}{\intercap}

\newcommand{\expose}{\mbox{ExPose}\xspace}

\newcommand{\prox}{\mbox{\mbox{PROX}}\xspace}
\newcommand{\proxD}{\mbox{\mbox{PROX-D}}\xspace}

\newcommand{\phosa}{\mbox{\mbox{PHOSA}}\xspace}
\newcommand{\artec}{\mbox{\mbox{Artec}}\xspace}
\newcommand{\rich}{\mbox{\mbox{RICH}}\xspace}

\newcommand{\vibe}{\mbox{VIBE}\xspace}

\newcommand{\ourTitle}{\ourname: Joint Markerless 3D Tracking \\ of Humans and Objects in Interaction}

\newcommand{\behave}{\mbox{BEHAVE}\xspace}
\newcommand{\BEHAVE}{\behave}
\newcommand{\mover}{\mbox{MOVER}\xspace}
\newcommand{\MOVER}{\mover}

\newcommand{\pigraphs}{{\mbox{PiGraphs}}\xspace}
\newcommand{\grab}{{\mbox{GRAB}}\xspace}

\newcommand{\smplifyx}{\mbox{SMPLify-X}\xspace}
\newcommand{\smplifyX}{\smplifyx}

\newcommand{\smpl}{\mbox{SMPL}\xspace}

\newcommand{\smplx}{\mbox{SMPL-X}\xspace}
\newcommand{\smplX}{\smplx}

\newcommand{\vposer}{\mbox{VPoser}\xspace}

\newcommand{\openpose}{\mbox{OpenPose}\xspace}

\newcommand{\mocap}{\mbox{MoCap}\xspace}
\newcommand{\vicon}{\mbox{Vicon}\xspace}

\newcommand{\twoD}{2D\xspace}
\newcommand{\threeD}{3D\xspace}
\newcommand{\fourD}{4D\xspace}

\newcommand{\mExpr}{\mathcal{E}}
\newcommand{\mColl}{\mathcal{P}}
\newcommand{\mContact}{\mathcal{C}}

\newcommand{\threedpw}{\mbox{3DPW}\xspace}
\newcommand{\threeDPW}{\threedpw}

\newcommand{\etal}{\mbox{et al.}\xspace}
\newcommand{\ie}{\mbox{i.e.}\xspace}
\newcommand{\eg}{\mbox{e.g.}\xspace}

\newcommand{\rgb}{\mbox{RGB}\xspace}
\newcommand{\rgbD}{\mbox{RGB-D}\xspace}

\newcommand{\includegraphicstrimW}[3]{%
  \begingroup
    \edef\x{\endgroup\noexpand\includegraphics[trim={#2}, clip=true, width=#3]{#1}%
    }%
   \x%
}

\newcommand{\CRColor}{\textcolor{black}}
\newcommand{\CRColorSUPMAT}{\textcolor{black}}

\begin{document}
\def\SubNumber{10}

\def\GCPRTrack{Main Track}

\title{\ourTitle}
\author{
    Yinghao Huang\inst{1}       \and 
    Omid Taheri\inst{1}         \and 
    Michael J. Black\inst{1}    \and 
    Dimitrios Tzionas\inst{2}}
\authorrunning{Huang et al.}
\institute{
    $^1$ Max Planck Institute for Intelligent Systems, T\"{u}bingen, Germany \\
    $^2$ University of Amsterdam, Amsterdam, The Netherlands \\
\email{\{yhuang2, otaheri, black\}@tue.mpg.de, d.tzionas@uva.nl}}

\maketitle
\begin{abstract}
Humans constantly interact with daily objects to accomplish tasks. To understand such interactions, \CRColor{computers need} to reconstruct these from cameras observing \CRColor{whole-body} interaction with scenes. \CRColor{This is challenging} due to occlusion between the body and objects, motion blur, depth/scale ambiguities, and the low image resolution of hands and graspable object parts. To make the problem tractable, the community focuses either on interacting hands, ignoring the body, or on interacting bodies, ignoring hands. The \grab dataset addresses dexterous whole-body interaction but uses \CRColor{marker-based}~\mocap and lacks images, while \mbox{BEHAVE} captures video of body-object interaction but lacks hand detail. We address the limitations of prior work with \ourname, a novel method that reconstructs \CRColor{interacting whole-bodies and objects} from \CRColor{multi-view}  \mbox{RGB-D} data, using the parametric \CRColor{whole-body} model \smplX and known object meshes. To tackle the above challenges, \ourname uses two key observations: (i) Contact between the hand and object can be used to improve the pose estimation of both. (ii) \kinectAZ sensors allow us to set up a simple multi-view \rgbD capture system that minimizes the effect of occlusion while providing reasonable inter-camera synchronization. With this method we capture the \ourname dataset, which contains \ourNumbSubj subjects (\ourNumbSubjM males and \ourNumbSubjM females) interacting with \ourNumbObj objects of various sizes and affordances, including contact with the hands or feet. In total, \ourname has \ourNumbMotions \rgbD videos, resulting in \ourNumbFrames multi-view frames, each containing \ourNumbCam~\rgbD images. Our method provides pseudo ground-truth body meshes and objects for each video frame. Our \ourname method and dataset fill an important gap in the literature and support many research directions. Our data and code are available for research purposes at \websiteURL. 
\end{abstract}
         \enlargethispage{1.0 \baselineskip}

\begin{figure}
    \centering
    \includegraphics[trim=000mm 000mm 000mm 000mm, clip=false, width=1.00\textwidth, keepaspectratio]{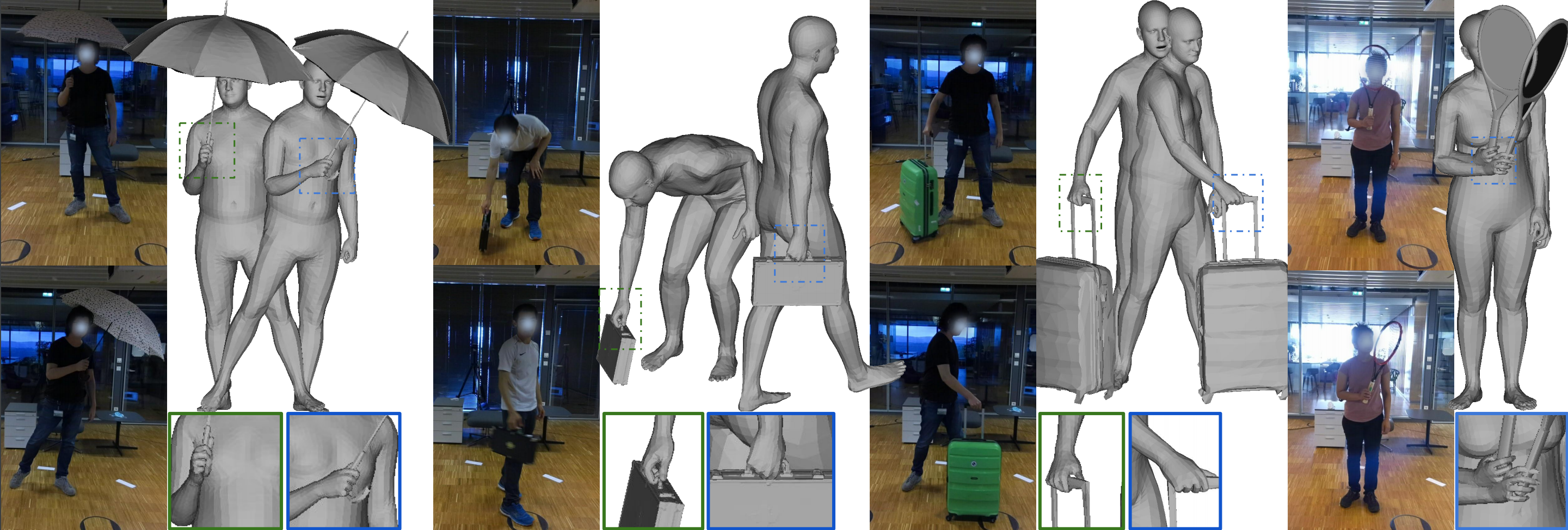}
    \caption{
    Humans interact with objects to accomplish tasks. To understand such interactions we need the tools to reconstruct them from whole-body videos in \fourD, \ie, as \threeD meshes in motion. Existing methods struggle, due to the strong occlusions, motion blur, and low-resolution of hands and object structures in such videos. Moreover, they mostly focus on the main body, ignoring the hands and objects. We develop \ourname, a novel method that reconstructs plausible interacting whole-body and object meshes from multi-view \rgbD videos, using contact constraints to account for strong ambiguities. With this we capture the rich \ourname dataset of \ourNumbMotions \rgbD videos (\ourNumbFrames multi-view frames, with \ourNumbCam~\kinectAZs) containing \ourNumbSubj subjects (\ourNumbSubjM fe-/males) interacting with \ourNumbObj objects of various sizes and affordances; note the hand-object grasps. 
    }
    \label{fig:teaser}
\end{figure}

\section{Introduction}   \label{sec:intro}

A long-standing goal of Computer Vision is to understand human actions from videos. Given a \CRColor{video} people effortlessly figure out what objects exist in it, the spatial layout of objects, and the pose of humans. Moreover, they deeply understand the depicted action. What is the subject doing? Why  are they doing this? What is their goal? How  do they achieve this? To empower computers with the ability to infer such abstract concepts from pixels, we need to capture rich datasets and to devise appropriate algorithms.

Since humans live in a \threeD world, their physical actions involve interacting with objects. \CRColor{Think of} how many times per day one goes to the kitchen, grabs a cup of water, and drinks from it. This involves contacting the floor with the feet, contacting the cup with the hand, moving the hand and cup together while maintaining contact, and drinking while the mouth contacts the cup. Thus, to understand human actions, it is necessary to reason in \threeD about humans and objects \emph{jointly}. 

There is significant prior work on estimating~\threeD humans without taking into account \CRColor{objects} \cite{bogo2016keep} and estimating~\threeD \CRColor{objects} without taking into account humans \cite{zollhofer2018state}.
There is even recent work on inserting bodies into \threeD scenes such that their interactions appear realistic \cite{zhang2020generating, li2019putting, Hassan2021posa}. But there is little work on estimating \threeD humans interacting with scenes \GCPR{and moving objects}, in which the human-scene/object contact is explicitly modeled and exploited. To study this problem, we need a dataset of videos with rich human-object interactions and reliable \GCPR{\threeD} ground truth.

\prox~\cite{hassan2019resolving} takes a step in this direction by estimating the \threeD body in a known \threeD scene. The scene mesh provides information that helps resolve pose ambiguities commonly encountered when a single camera is used. However, \prox involves only coarse interactions of bodies, static scenes \CRColor{with no moving objects}, and \CRColor{no dexterous fingers.} The recent \BEHAVE dataset \cite{bhatnagar2022behave} uses multi-view \rgb-D data to capture humans interacting with objects but does not include detailed hand pose or fine hand-object contact. Finally, the \grab dataset \cite{taheri2020grab} captures the kind of detailed hand-object and whole-body-object interaction that we seek but is captured using marker-based \mocap and, hence, lacks images paired with the ground-truth scene.

We argue that what is needed is a new dataset of \GCPR{\rgb} videos containing natural human-object interaction in which the \CRColor{whole} body is tracked reliably, the hand pose is captured, objects are also tracked, and the hand-object contact is realistic; see \fig{fig:teaser}. This is challenging and requires technical innovation to create. To that end, we design a system that uses multiple \rgbD sensors that are spatially calibrated and temporally synchronized. To this data we fit the \smplX body model, which has articulated hands, by extending the \prox~\cite{hassan2019resolving} method to \CRColor{use} multi-view data \CRColor{and  grasping hand-pose priors}. We also track the \threeD objects with which the person interacts. The objects used in this work are representative of  items one finds in daily life. We obtain \GCPR{accurate}~\threeD models for each \CRColor{object}~\GCPR{with a handheld \artec scanner}. Altogether we collect \ourNumbMotions sequences (\ourNumbFrames multi-view frames), with \ourNumbSubj subjects interacting with \ourNumbObj objects.

The problem, however, is that separately estimating the body and objects is not sufficient to ensure accurate \threeD body-object contact. Consequently, a key innovation of this work is to estimate these \emph{jointly}, while exploiting information about \emph{contact}. Objects do not move independently, so, when they \CRColor{move}, it means the body is in contact. We define likely contact regions on objects and on the body. Then, given frames with known \GCPR{likely} contacts, we enforce contact between the body and the object when estimating the body and object poses. The resulting method produces \CRColor{natural} body poses, hand poses, and object poses. Uniquely, it provides detailed \GCPR{pseudo} ground-truth contact information between the \CRColor{whole} body and objects in \GCPR{\rgb} video.

In summary, our major contributions are as follows:
    (1)     We develop a novel Motion Capture \CRColor{method} utilizing multiple \rgbD cameras. It is relatively lightweight and flexible, yet accurate enough, thus suitable for data capture of daily scenarios. 
    (2) We extend previous work on fitting \smplX to images to fit it to multi-view \rgbD data while taking into account body-object contact.
    (3)     We capture a novel dataset that contains whole-body human motions and interaction with objects, as well as multi-view \rgbD imagery. Our data and code are available at \websiteURL.

\section{Related Work}   \label{sec:related}

There is a large literature 
on estimating \threeD human pose and shape 
from images or videos \cite{bogo2016keep, pavlakos2019expressive, choutas2020monocular, kanazawa2018end, kocabas2020vibe, varol2017long, mehta2017vnect, omran2018neural}.
Here we focus on the work most closely related to ours, particularly as it concerns, or enables, capturing human-object interaction. 

\textbf{\mocap from Multi-view Videos and \imus.}
Markerless \mocap from multi-view videos \cite{liu2011markerless, de2008performance, huang2017towards} is widely studied and commercial solutions exist (\eg,~Theia Markerless).
Compared with traditional marker-based \mocap, markerless offers advantages of convenience,  applicability in outdoor environments, \CRColor{non-intrusiveness}, and greater flexibility. 
However, traditional \mocap methods, both marker-based and markerless, focus on extracting a \threeD skeleton.
This is useful for biomechanics but our goal is to reason about body-scene contact.
To enable that, we need to capture the body surface.

Various \threeD human representations have been proposed, with recent work focused on learning a parametric \CRColor{mesh-based} model of body shape from large-scale collections of \threeD scans \cite{anguelov2005scape, loper2015smpl, romero2017embodied, pavlakos2019expressive, osman2020star, xu2020ghum, supr2022}. 
Here we use the \smplx model \cite{pavlakos2019expressive} because it contains fully articulated hands, which are critical for reasoning about object manipulation.
The  body parameters are often estimated by fitting the \threeD generative model to various \twoD cues like landmarks detected by Convolutional Neural Networks \cite{cao2019openpose, wei2016convolutional, newell2016stacked} or silhouettes \cite{rhodin2016general, xu2018monoperfcap, alldieck2018video}. 
Though effective, these monocular video-based methods suffer from depth ambiguity and occlusions. 
To address this issue, researchers have proposed to combine \imus with videos to obtain better and more robust results \cite{von2018recovering, pons2010multisensor}. 

Many methods estimate \threeD bodies from multi-view images but focus on skeletons and not \threeD bodies
\cite{epipolartransformers2020cvpr,iskakov2019learnable,Qiu_2019_ICCV,to2020voxelpose,Dong_2019_CVPR,Dong_2021_PAMI,Zhang_2020_CVPR}.
Recent work addresses \threeD body shape estimation from multiple views \cite{huang2017towards,dong2021shape,lightcap2021}.
Most related to our work are two recent datasets. 
The \rich dataset \cite{Huang:CVPR:2022}, fits \smplX bodies to multi-view \rgb videos taken both indoors and outdoors.
The method uses a detailed \threeD scan of the scene and models the contact between the body and the world.
\rich does not include any object motion; the scenes are completely rigid.
In contrast, \BEHAVE~\cite{bhatnagar2022behave} contains \smpl bodies interacting with \threeD objects that move. 
We go beyond that work, however, to integrate novel contact constraints and to capture hand pose, which is critical for human-object interaction.
Additionally, \BEHAVE focuses on large objects like boxes and chairs, whereas we have a wider range of object sizes, including smaller objects like cups.

\textbf{Human-Object Interaction.}
There has been a lot of work on modeling or analyzing human-object interactions \cite{yao2010modeling, Hamer_Hand_Manipulating_2009, oikonomidis2011full, Rogez2015everyday, tzionas2016capturing, hampali2020honnotate, hasson2019learning, karunratanakul2020grasping, bhatnagar2022behave}. A detailed discussion is out of the scope of this work. Here, we focus on modeling and analyzing human-object interaction in \threeD space. Most existing work, however, only focuses on estimating hand pose \cite{hasson2019learning, hampali2020honnotate, hasson2020leveraging, romero2010handsInAction}, ignoring the strong relationship between body motion, hand motion, and object motion. Recent work considers whole-body motion. For example, the \grab~\cite{taheri2020grab} dataset  provides detailed object motion and whole-body motion in a parametric body format (\smplX). Unfortunately, it is based on \mocap and does not include video. Here our focus is on tracking the \CRColor{whole}-body \GCPR{motion}, object motion, and the detailed hand-object contact to provide ground-truth \threeD information in \rgb video.

\textbf{Joint Modeling of Humans and Scenes.}
There is some prior work addressing human-object contact in both static images and video. For example, \phosa estimates a \threeD body and a \threeD object with plausible interaction from a single \rgb image \cite{zhang2020phosa}. Our focus here, however, is on dynamic scenes. Motivated by the observation that natural human motions always happen inside \threeD scenes, researchers have proposed to model human motion jointly with the surrounding environment \cite{hassan2019resolving, savva2016pigraphs, cao2020long, yi2022mover}. In \prox~ \cite{hassan2019resolving} the contact between humans and scenes is explicitly used to resolve ambiguities in pose estimation. The approach avoids bodies interpenetrating scenes while encouraging contact between the scene and nearby body parts. Prior work also tries to infer the most plausible position and pose of humans given the \threeD scene \cite{zhang2020generating, li2019putting, Hassan2021posa}. Most recently, \MOVER~\cite{yi2022mover} estimates the \threeD scene and the \threeD human directly from a static monocular video in which a person interacts with the scene. While the \threeD scene is ambiguous and the human motion is ambiguous, by exploiting contact, the method resolves many ambiguities, improving the estimates of both the scene and the person. Unfortunately, this assumes a static scene and does not model hand-object manipulation.

\textbf{Datasets.}
Traditionally, \mocap is performed using marker-based systems inside lab environments. To capture object interaction and contact, one approach uses MoSh \cite{loper2014mosh} to fit a \smpl or \smplX body to the markers \cite{mahmood2019amass}. An advanced version of this is used for \grab~\cite{taheri2020grab}. Such approaches lack synchronized \rgb video. The HumanEva \cite{sigal2006humaneva} and Human3.6M \cite{ionescu2013human3} datasets combine multi-camera \rgb video capture with synchronized ground-truth \threeD skeletons from marker-based \mocap. These datasets lack ground-truth \threeD body meshes, are captured in a lab setting, and do not contain human-object manipulation. \threeDPW~\cite{von2018recovering} is the first in-the-wild dataset that jointly features natural human appearance in video and accurate \threeD pose. This dataset does not track objects or label human-object interaction. \pigraphs~\cite{savva2016pigraphs} and \prox \cite{hassan2019resolving}  provide both \threeD scenes and human motions but are relatively inaccurate, relying on a single \rgbD camera. This makes these datasets ill-suited as evaluation benchmarks. The recent \rich dataset \cite{Huang:CVPR:2022} addresses many of these issues with indoor and outdoor scenes, accurate multi-view capture of \smplX, \threeD scene scans, and human-scene contact. It is not appropriate for our task, however, as it does not include object manipulation.
\begin{wraptable}{r}{0.55\linewidth}
    \scriptsize
    \centering
    \vspace{-1.6 em}
    \begin{tabular}{|l|c|c|c|c|c|c|}                                                         \hline
        Name                & \# of & Natural   &  Moving   & Accurate  & With    & Artic.  \\
                            &  Seq.  & Appear.   &  Objects  & Motion    & Image  & Hands \\  \hline
        HumanEva~\cite{sigal2006humaneva}     & 56      & \cmark    & \xmark    & \cmark    & \cmark  &  \xmark \\
        Human3.6M~\cite{ionescu2013human3}           & 165     & \cmark    & \xmark    & \cmark    & \cmark  &  \xmark  \\
        AMASS~\cite{mahmood2019amass}               & 11265   & \cmark    & \xmark    & \cmark    & \xmark  &  \xmark \\
        GRAB~\cite{taheri2020grab}                & 1334    & \cmark    & \cmark    & \cmark    & \xmark  &  \cmark \\
        3DPW~\cite{von2018recovering}           & 60      & \cmark    & \xmark    & \cmark    & \cmark  &  \xmark \\  \hline
        GTA-IM~\cite{cao2020long}              & 119     & \xmark    & \xmark    & \cmark    & \cmark  &  \xmark \\
        SAIL-VOS~\cite{hu2019sail}            & 201     & \xmark    & \xmark    & \xmark    & \xmark  &  \xmark \\  \hline
        \pigraphs~\cite{savva2016pigraphs}           & 63      & \cmark    & \xmark    & \cmark    & \cmark  &  \xmark \\
        \prox~\cite{hassan2019resolving}               & 20      & \cmark    & \xmark    & \xmark    & \cmark  &  \xmark \\
        RICH~\cite{Huang:CVPR:2022}              & 142     & \cmark    & \xmark    & \cmark      & \cmark  &  \xmark \\
        BEHAVE~\cite{bhatnagar2022behave}              & 321     & \cmark    & \cmark    & \cmark    & \cmark  &  \xmark \\
        \textbf{\ourname} (ours)   & \ourNumbMotions     & \cmark    & \cmark    & \cmark  & \cmark & \cmark  \\  
        \hline
    \end{tabular}
    \vspace{-1.0 em}
    \caption{
                Dataset statistics.
                Comparison of our \ourname dataset to existing datasets.
    }
    \label{tab:dataset_compare}
    \vspace{-2.0 em}
\end{wraptable} An alternative approach is the one of \mbox{GTA-IM} \cite{cao2020long} and \mbox{SAIL-VOS} \cite{hu2019sail}, which generate human-scene interaction data using either \threeD graphics or \twoD videos. They feature high-accuracy ground truth but lack visual realism. In summary, we believe that a \threeD human-object interaction dataset needs to have accurate hand poses to be useful, since  hands are how people most often interact with objects. We compare our \ourname dataset with other ones in \tab{tab:dataset_compare}.

\section{\CRColor{\ourname} Method}      \label{sec:method}

Our core goal is to accurately estimate the human \CRColor{and object} motion throughout a video. Our markerless motion capture method is built on top of the \proxD method \CRColor{of Hassan~\etal}~\cite{hassan2019resolving}. To improve the body tracking accuracy we extend this method to use multiple RGB-D cameras; \CRColor{here we use} the latest \kinectAZ cameras. The motivation is that multiple cameras observing the body from different angles give more information about the human and object motion. Moreover, \GCPR{commodity}~\rgbD cameras are much more flexible to deploy out of controlled lab scenarios than more specialized devices. 

The key technical challenge lies in accurately estimating  the \threeD pose and translation of the objects while a person interacts with them. In this work we focus on $10$ variously sized rigid objects common in daily life, \CRColor{such as} cups and chairs. Being rigid does not make the tracking of the objects trivial because of the occlusion by the body and hands. While there is a rich literature on 6 DoF object pose estimation, much of it ignores hand-object interaction. Recent work in this direction is promising but still focuses on scenarios that are significantly simpler than ours, cf.~\cite{sun2022onepose}.

\GCPR{Similar} 
to previous work on hand and object pose estimation \cite{hampali2020honnotate} from \rgbD videos, in this work we assume that the \threeD meshes of the objects are known in advance. To this end, we first gather the \threeD models of these objects from the Internet whenever possible and  scan the remaining objects ourselves. To fit the known object models to image data, we first preform semantic segmentation, find the corresponding object regions in all camera views, and fit the \threeD mesh to the segmented object contours via differentiable rendering. Since heavy occlusion between humans and objects in some views may make the segmentation results unreliable, aggregating segmentation from all views boosts the object tracking performance. 

In the steps above, both the subject and object are treated separately and processing is per frame, with no temporal smoothness or contact constraint applied. This produces jittery motions and heavy penetration between objects and the body. Making matters worse, our human pose estimation exploits 
\openpose for 2D keypoint detection, which \CRColor{struggles} when the object occludes the body or the hands interact with it. To mitigate this issue and still get reasonable \CRColor{body,} hand and object pose in these challenging cases, we manually annotate the frames where the \CRColor{body or the hand} is in contact with the object, \CRColor{as well as} the \CRColor{body,} hand and object vertices that are most likely to be in contact. \CRColor{This manual annotation can be tedious; automatic detection of contact is an open problem and is left for future work.} We then explicitly encourage the labeled \CRColor{body and} hand vertices to be in contact with the labeled object vertices. We find that this straightforward idea works well in practice. More details are described in the following. 

\subsection{Multi-Kinect Setup}

We use 6 \kinectAZs to track the human and object \GCPR{together}, deployed \CRColor{in a ``ring'' layout} in an office; see \supmat Multiple \rgbD cameras provide a good balance between body tracking accuracy and applicability to real scenarios, compared with costly professional \mocap systems \GCPR{like}~\vicon, or cheap and convenient but not-so-accurate monocular \rgb cameras. Moreover, this approach does not require \CRColor{applying any} markers, making the images natural. \GCPR{Intrinsic camera parameters are provided by the manufacturer.} \CRColor{Extrinsic camera parameters are obtained via} camera calibration with \kinectAZ's API \cite{kinectAPI}. \CRColor{However, these can be a bit noisy, as non-neighbouring cameras in a sparse ``ring'' layout don't observe the calibration board well at the same time. Thus, we manually refine in MeshLab the extrinsics by comparing the point clouds for neighbouring cameras for several iterations.} \CRColor{The hardware synchronization of \kinectAZs is empirically reasonable.} Given \GCPR{the calibration} information, we choose one camera's \threeD coordinate frame as the global frame and transform the point clouds from the other frames into the global frame, which is where we fit the \smplX \CRColor{and object models}.

\subsection{Sequential Object-Only Tracking}

\textbf{Object Segmentation.}
To track an object during interaction, we need reliable visual cues about it to compare with the \threeD object model. 
To this end,
we perform semantic segmentation by applying \pointrend~\cite{kirillov2020pointrend} to the whole image. 
We then extract the object instances that correspond to the categories of our objects; 
\GCPR{for examples see \supmat} 
We assume 
that the subject interacts with a single object. 
Note that, in contrast to previous approaches where the objects occupy a large portion of the image \cite{hampali2020honnotate,hassan2019resolving,tzionas2016capturing,oikonomidis2011full}, in our case 
\GCPR{the entire body is visible, 
thus, the object takes up a small part of the image and is often occluded by the body and hands; our setting is much more challenging.}
We observe that \pointrend works reasonably well for large objects like chairs, even with heavy occlusion between the object and the human, while for small objects, like a bottle or a cup, 
\CRColor{it struggles} significantly due to 
occlusion.

In extreme cases, it is possible for the object to 
\GCPR{not be detected in} 
most of the views. 
But 
even when the segmentation is good, the class label for the objects may be wrong. 

\GCPR{To resolve this, we take two steps:} 
\GCPR{(1)
For every frame, we \CRColor{detect} 
all possible object segmentation candidates and their labels}. This step takes place offline and only once. 
\GCPR{(2)
During the object tracking phase, 
for each view, we compare 
the rendering of the tracked object from the $i^{\mathrm{th}}$ frame with 
all the detected segmentation candidates for the \GCPR{$(i+1)^{\mathrm{th}}$ frame}, and preserve
only the candidate with the largest overlap ratio.} 
This 
\CRColor{render-compare-and-preserve} 
operation takes place \CRColor{iteratively} during tracking.

\textbf{Object Tracking.}
Given object 
\GCPR{masks} 
via semantic segmentation over the whole sequence, we track the object 
\GCPR{by fitting 
its model} to 
observations via differentiable rendering \cite{kato2018neural, loper2014opendr}. 
This
is similar to past work for hand-object tracking \cite{hampali2020honnotate}. 
We
assume that the object is rigid and its mesh is given. 
The configuration of the rigid object in the $t^{\mathrm{th}}$ frame 
is 
specified via a 6D rotation and translation vector $\xi$. 
\CRColor{For initialization, we manually obtain the configuration of the object for the first frame by matching the object mesh into the measured point clouds.} Let \GCPR{$R_S$ and $R_D$ be functions that render a synthetic mask and depth image for the tracked \threeD object mesh, $M$}. Let also $S = \{S_{\nu}\}$ be the \GCPR{``observed''} object masks and $D = \{D_{\nu}\}$ be corresponding depth values for the current frame, \GCPR{where $\nu$ is the camera view.} Then, we minimize:
{\small
\begin{align}
\begin{split}
E_O(\xi; S, D) = 
\sum_{{\text{view}~\nu}} ~ 
& \lambda_{segm}  \| (R_S(\xi, M, {\nu}) - S_{\nu}) * S_{\nu} \|_F^2 +    \\
& \lambda_{depth} \| (R_D(\xi, M, {\nu}) - D_{\nu}) * S_{\nu} \|_F^2    \text{,}
\label{eq:fit_object}
\end{split}
\end{align}
}
\GCPR{where the two terms compute how well the rendered object mask and depth \CRColor{image} match the detected \CRColor{mask and observed depth;} the $*$ is an element-wise multiplication, and $\|.\|_F$ the Frobenius norm}; \CRColor{ $\lambda_{segm}$ and $\lambda_{depth}$ are steering weights
\CRColor{set empirically}. }For simplicity, we assume that transformations from the master to other camera frames are encoded in the rendering functions $R_S, R_D$; we do not denote these explicitly here.

\subsection{Sequential Human-Only Tracking} 

We estimate body shape and pose over the whole sequence from multi-view \rgbD videos in a frame-wise manner. This is similar in spirit with the \proxD method \cite{hassan2019resolving}, but, in our case, there is no \threeD scene constraint and multiple cameras are used. The human pose and shape are optimized independently in each frame. We use the \smplX~\cite{pavlakos2019expressive} model to represent the \threeD human body. \smplX is a function that returns a water-tight mesh given parameters for shape, $\beta$, pose, $\theta$, facial expression, $\psi$, and translation, $\gamma$. We follow the common practice of \CRColor{using a $10$-dimensional space for} shape, $\beta$, and a $32$-dimensional latent space in \vposer \cite{pavlakos2019expressive} to present body pose, $\theta$.

We minimize the loss defined below. For \CRColor{each} frame we essentially extend the major loss terms used in \prox~\cite{hassan2019resolving} to multiple views:
{\small
\begin{align}
\begin{split}
    E_B(\beta, \theta, \psi, \gamma; K, J_{est}) = E_J + & \lambda_D E_D + \lambda_{\theta_b} E_{\theta_b} + 
    \lambda_{\theta_h} E_{\theta_h} +     
    \lambda_{\theta_f} E_{\theta_f} +     
    \\
   &\lambda_\alpha E_\alpha + \lambda_\beta E_\beta + \lambda_\mExpr E_\mExpr + 
    \lambda_\mColl E_\mColl       \text{,} 
\label{eq:fit_body}
\end{split}
\end{align}
}%
where 
$E_\beta$, $E_{\theta_b}$, $E_{\theta_h}$, $E_{\theta_f}$, $ E_\mExpr$ are prior loss terms for body shape, body pose, hand pose, facial pose and expressions. Also, $E_\alpha$ is a prior for extreme elbow and knee bending. \GCPR{For detailed definitions of these terms see \cite{hassan2019resolving}.} $E_J$ is a  \twoD keypoint \GCPR{re-projection} loss: 
{
\small
\begin{equation}
    E_J(\beta, \theta, \gamma; K, J_\text{est} ) = 
        \sum_{\mathit{view}~\nu} ~ \sum_{\mathit{joint}~i} k_i^\nu w_i^\nu \rho_J
        \big(   \mathit{\Pi}_K^\nu  \big(   R_{\theta \gamma}   (   J(\beta)_i  
                                                                )   
                                    \big)   - J_{est, i}^\nu    
        \big)  \text{,}
\end{equation} 
}
\REBT{where \CRColor{$\theta = \{  \theta_b, \theta_h, \theta_f \}$,} $\nu$ and $i$ iterate through views and joints, $k_i^\nu$ and $w_i^\nu$ are the per-joint weight and detection confidence, $\rho_J$ is \GCPR{a robust Geman-McClure error function \cite{GemanMcClure1987}}, $\mathit{\Pi}_K^\nu$ is the projection function \GCPR{with $K$ camera parameters}, \GCPR{$R_{\theta \gamma}(J(\beta)_i)$ are the posed \threeD joints of \smplX,} and $J_{est, i}^\nu$ the detected \twoD joints}. The term $E_D$ is: 
{
\small
\begin{equation}
E_D(\beta, \theta, \gamma; K) = \sum_{\mathit{view}~\nu}    ~ 
                                    \sum_{p \in P^{\nu}}        ~
                                    \min_{v \in V_b^{\nu}} \|v-p\|    
                                    \text{,}
\end{equation}
} \GCPR{where $P^{\nu}$       is \kinectAZ's segmented point cloud for the $\nu^\text{th}$ view, and 
$V_b^{\nu}$     are \smplX vertices that are visible in this view}. \GCPR{This term measures how far the estimated body mesh is from the combined point clouds, \CRColor{so that we minimize this discrepancy}. Note that, unlike \prox, we have multiple point clouds from all views, \ie, our $E_D$ is a multi-view extension of \prox's~\cite{hassan2019resolving} loss}. For each view we dynamically compute the visible body vertices, and ``compare'' them against the segmented point cloud for that view. 

\GCPR{Finally, the term} $ E_\mColl $ penalizes self-interpenetration of the \smplX body mesh; see \prox~\cite{hassan2019resolving} for a more detailed and formal definition of this:
\begin{equation}
    E_\mColl(\theta, \beta, \gamma) = E_{\mColl_{self}}(\theta, \beta). 
    \label{eq:penetration}
\end{equation}
       
\subsection{Joint Human-Object Tracking Over All Frames}

\begin{figure}[t]
    \centering
    \includegraphics[height=0.186 \linewidth]{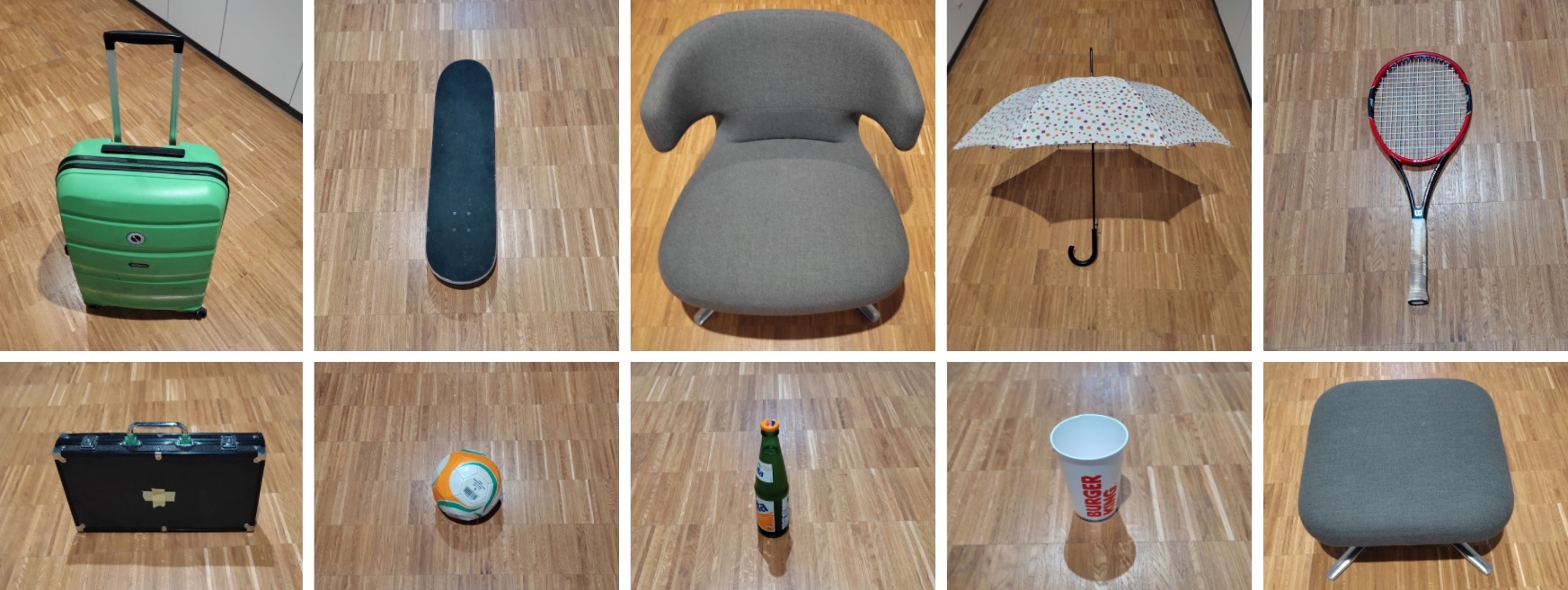} ~\vrule~
    \includegraphics[height=0.186 \linewidth]{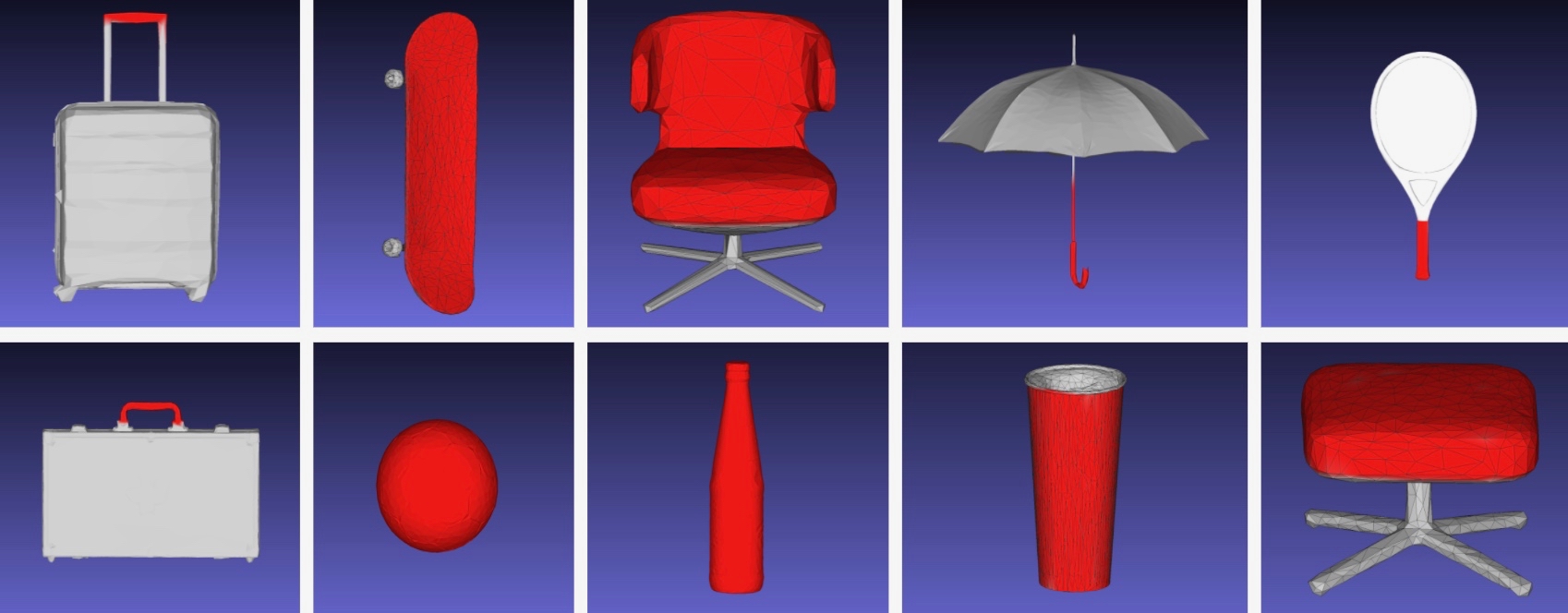}
    \caption{ The objects of our \ourname dataset. \textbf{Left:} Color photos. \textbf{Right:} Annotations for object areas that are likely to come in contact during interaction, shown in red.}
    \label{fig:object_photos_annotations}
\end{figure}

\newcommand{\FigThreeCustomWidth}{0.31}
\begin{wrapfigure}{r}{\FigThreeCustomWidth\textwidth}
    \vspace{-3.8 em}
    \begin{center}
    \includegraphics[trim={10mm 22mm 15mm 40mm}, clip=True, 
    width=\FigThreeCustomWidth\textwidth]{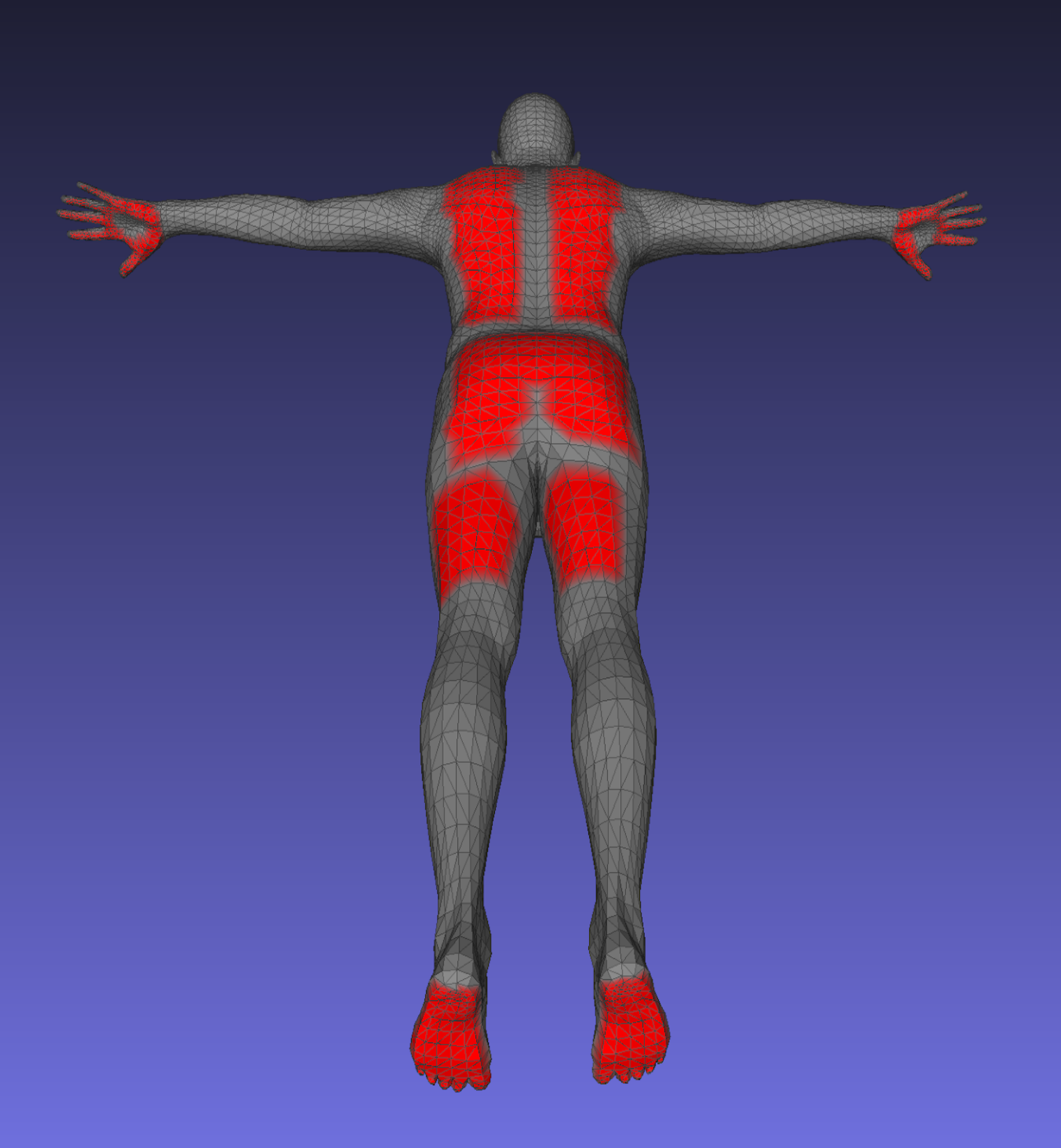}   
    \end{center}
    \vspace{-2.0 em}
    \caption{Annotation of likely body contact areas (red color).}
    \vspace{-3.0 em}
    \label{fig:body_annotations}
\end{wrapfigure}

We treat the result of the above optimization as initialization for refinement via \emph{joint} optimization of the body and the object \emph{over all frames}, subject to \emph{contact} constraints.

For this we fix the body shape parameters, \CRColor{$\beta$,} as the mean body shape computed over all frames from the first stage, as done in \cite{huang2017towards}. Then, we jointly optimize the object pose and translation, \CRColor{$\xi$}, body pose, \CRColor{$\theta$}, and body translation, \CRColor{$\gamma$}, over all frames. We add a temporal smoothness loss to reduce jitter for both the human and the object. We also penalize the \REBT{body-object} interpenetration, as done in \prox~ \cite{hassan2019resolving}. A key difference is that in \prox the scene is static, while here the object is free to move. 

To enforce contact, we annotate the \CRColor{body areas} that are most likely to be in contact with the objects and, for each object, we label vertices most likely to be contacted. These annotations are shown in \fig{fig:body_annotations} and \fig{fig:object_photos_annotations}\ourrefcolor{-right}, respectively, in red. We also annotate frame sub-sequences where \CRColor{the body is} in contact with objects, and enforce contact between them explicitly to get reasonable tracking even when there is heavy interaction and occlusion between hands and objects. Such interactions prove to be challenging for state-of-the-art \twoD joint detectors, \eg, \openpose, \CRColor{especially for hands}. 

Formally, \GCPR{we perform global optimization over all $T$ frames, and minimize} a loss, $E$, that is composed of an object fitting loss, $E_O$, a body fitting loss, $E_B$, \REBT{a motion smoothness prior \cite{zhang2021learning} loss, $E_{\mAcc}$,} and a loss penalizing object acceleration, $E_{\mAccc}$. We also use a ground support loss, $E_{\mathcal{G}}$, that \GCPR{encourages the human and the object} to be above the ground plane, \ie, \GCPR{to not penetrate it}. Last, we use a \CRColor{body}-object contact loss, $E_{\mContact}$, that attaches the \CRColor{body} to the object for frames with contact. The loss $E$ is defined as:
{
\footnotesize
\begin{align}
\begin{split}
E = ~~
\frac{1}{T} 
\sum_{\text{frame}~t}         
&\bigg[
E_O(\Xi_t; \mathcal{S}_{t}, \mathcal{D}_{t}) +
E_B(\beta^*, \Theta_t, \Psi_t, \Gamma_t; \mathcal{J}_{\mathit{est}}) 
\bigg] + \\
\frac{1}{T} 
\sum_{\text{frame}~t}       
&\bigg[
E_\mColl(\Theta_t, \beta^*, \Gamma_t) + 
E_{\mContact}(\beta^*, \Theta_t, \Psi_t, \Gamma_t, \Xi_t, M) 
\bigg]
+ \\ 
\frac{\lambda_{\mathcal{G}}}{T} 
\sum_{\text{frame}~t}         
&\bigg[
E_{\mathcal{G}}(\beta^*, \Theta_t, \Psi_t, \Gamma_t) + E_{\mathcal{G'}}(\Xi_t, M) 
\bigg]
+ \\
\frac{\lambda_\mathcal{Q}}{T} 
\sum_{\text{frame}~t}         
&\bigg[
Q_t * E_\mathcal{C}(\beta^*, \Theta_t, \Psi_t, M', \Xi_t) 
\bigg]      + \\
&
\lambda_\mAcc       E_\mAcc(\Theta, \Psi, \Gamma, A; \beta^*, T) + \\
&
\lambda_{\mAccc}    E_{\mAccc}(\Xi, T, M) 
\text{,}
\end{split}
\label{eq:big_loss}
\end{align}
}
where 
\GCPR{
$E_O$ comes from \ourrefcolor{Eq.~}\ref{eq:fit_object} and 
$E_B$       from \ourrefcolor{Eq.~}\ref{eq:fit_body}, and both go through all views $\nu$, 
\CRColor{while $E_\mColl$ comes from \ourrefcolor{Eq.~}\ref{eq:penetration}}.} 
\GCPR{For all frames $t = \{ 1, \dots, T\}$ of a sequence}, 
\GCPR{$\Theta = \{ \theta_t \}$}, 
\GCPR{$\Psi   = \{ \psi_t \}$}, 
\GCPR{$\Gamma = \{ \gamma_t \}$}, 
are the body poses, facial expressions and translations, 
\GCPR{$\Xi = \{ \xi_t \}$} is the object rotations and translations, 
\GCPR{$\mathcal{S} = \{ S_t \}$ and $\mathcal{D} = \{ D_t \}$ are masks and 
depth patches, 
and} 
\GCPR{$\mathcal{J}_{\mathit{est}} = \{ {J}_{\mathit{est}, t} \}$ are detected \twoD keypoints.}
$M$     is the object mesh, and 
\GCPR{$\beta^*$ the mean body shape}.
$E_\mathcal{C}$ encourages \CRColor{body}-object contact for frames in contact, which are indicated by the 
\GCPR{manually annotated binary vectors 
$Q = \{ 
\CRColor{Q_t}
\}$, \CRColor{$t=\{1, \dots, T \}$}; 
$Q_t$ is set to 1 if in the $t^\mathrm{th}$ frame \CRColor{any body part (\eg, hand, foot, thighs)} is in contact with the object, and set to 0 otherwise}. 
The motion smoothness loss $E_{\mathcal{S}}$ penalizes abrupt position changes for body vertices, and the vertex acceleration loss $E_{\mAccc}$ encourages smooth object trajectories. 
We estimate the ground plane surface by fitting a plane to chosen floor \CRColor{points in the observed point clouds}.
The terms $E_{\mathcal{G}}$ and $E_{\mathcal{G'}}$ measure whether the body and object 
\GCPR{penetrate the ground, respectively.} 
\CRColor{For more details on the above loss terms}, please see \supmat 
\CRColor{Finally, the parameters 
$\lambda_{\mathcal{G}}$, 
$\lambda_\mathcal{Q}$, 
$\lambda_\mAcc$, and 
$\lambda_{\mAccc}$
are steering weights that are set empirically.}

\section{\ourname Dataset}   \label{sec:dataset}

We use the proposed \ourname algorithm (\ourrefcolor{Sec.~}\ref{sec:method}) to capture the \ourname dataset, which uniquely features whole-body interactions with objects in multi-view \rgbD videos.
\medskip

\textbf{Data-capture Protocol.}
\REBT{We use \ourNumbObj \CRColor{everyday} objects, shown in \fig{fig:object_photos_annotations}\ourrefcolor{-left}, that vary in  size  and ``afford'' different interactions with the body, hands or feet}; \CRColor{we focus mainly on hand-object interactions}. We recruit \ourNumbSubj subjects (\ourNumbSubjM males and \ourNumbSubjF females) that are \CRColor{between} 25-40 years old. The subjects are recorded \CRColor{while} interacting with $7$ or more objects, \CRColor{according to their time availability}. Subjects are instructed to interact with objects as naturally as possible. However, \REBT{\CRColor{they are asked to} 
avoid very fast interactions that cause severe motion blur (\kinectAZ supports only up to $30$ FPS), \CRColor{or misalignment between the \rgb and depth images for each Kinect (due to technicalities of \rgbD sensors)}.} We capture up to 3 sequences per object \CRColor{depending on object shape and functionality, and by picking} an interaction intent from the list below, \CRColor{as in \grab~\cite{taheri2020grab}:} 
\begin{itemize}
    \item 
    "\textbf{Pass}": The subject \CRColor{passes} the object \CRColor{on} to another imaginary person \CRColor{standing on their left/right side; a graspable area needs to be free for the other person to grasp.}
    \item
    "\textbf{Check}": The subject \CRColor{inspects visually} the object \CRColor{from several viewpoints by first picking it up and then manipulating it with their hands to see several sides of it.}
    \item
    "\textbf{Use}": The subject \CRColor{uses} the object in a \CRColor{natural} way \CRColor{that ``agrees'' with the object's affordances and functionality for everyday tasks.}
\end{itemize}
We also capture each subject performing a freestyle interaction of their choice. All subjects gave informed written consent to \CRColor{publicly} share their data for research.

\begin{figure}[t]
    \centering
        \includegraphics[trim=000mm 000mm 000mm 000mm, clip=false, width=1.00\linewidth]{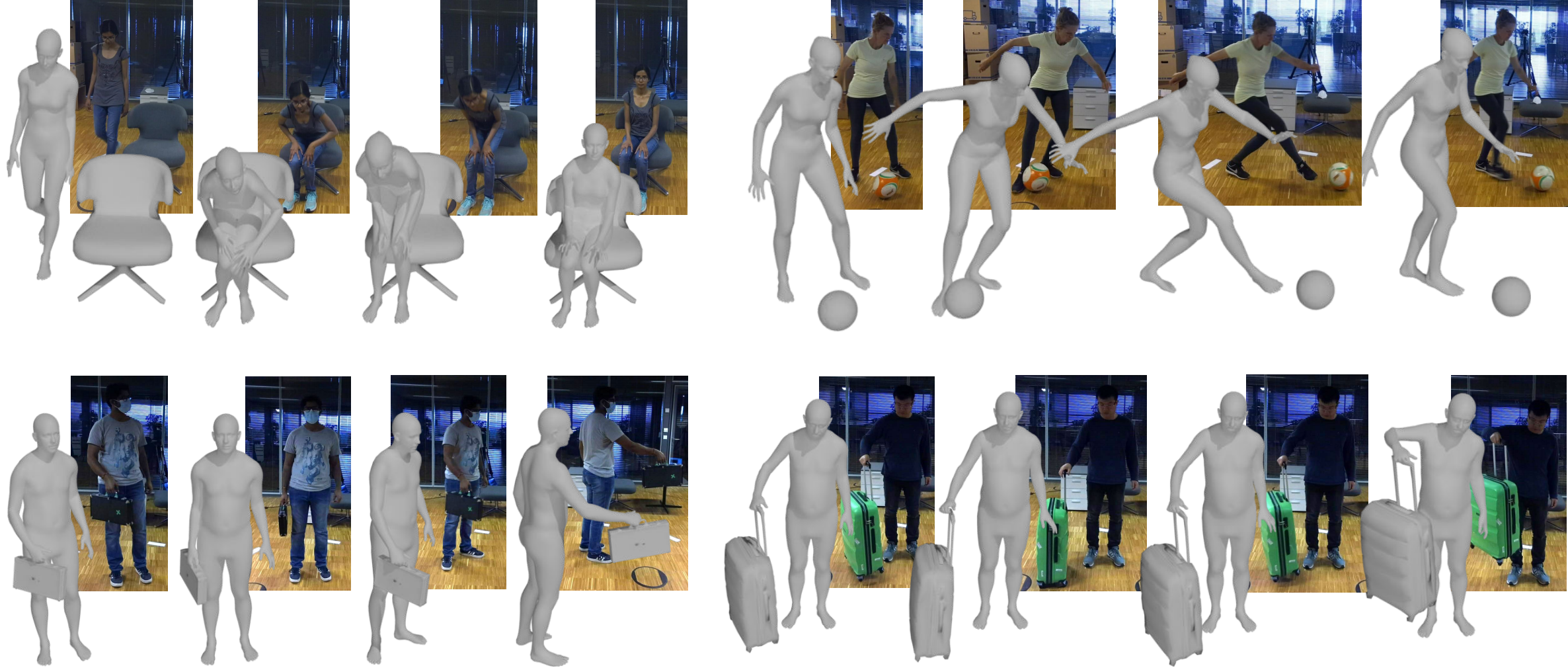}
    \caption{
                Samples from our \ourname dataset, drawn from four sequences with different subjects and objects. The estimated \threeD object and \smplX human meshes have plausible contacts that agree with the input images. \REBT{Best viewed zoomed in.}
    }
    \label{fig:dataset_sequences}
\end{figure}

\medskip

\textbf{4D Reconstruction.} Our \ourname method (\ourrefcolor{Sec.~}\ref{sec:method}) takes as input multi-view \rgbD videos and outputs \fourD meshes for the human and object, \ie, \threeD meshes over time. Humans are represented as \smplX meshes \cite{pavlakos2019expressive}, while object meshes are acquired with an Artec hand-held scanner. \REBT{Some dataset frames along with the reconstructed meshes are shown in \fig{fig:teaser} and \fig{fig:dataset_sequences}; see also \CRColor{the video on our website}.} Reconstructions look natural, with plausible contact between the human and the object. 

\medskip

\REBT{\textbf{Dataset Statistics.}} \ourname has \ourNumbMotions~\rgbD videos with a total of \ourNumbFrames multi-view frames (\ourNumbCam~\rgbD images each). \REBT{For a comparison with other datasets, see \tab{tab:dataset_compare}.}

\begin{figure}[t]
    \centering
    \includegraphics[height=0.236 \linewidth, trim={10mm 20mm 0mm 16mm}, clip=true]{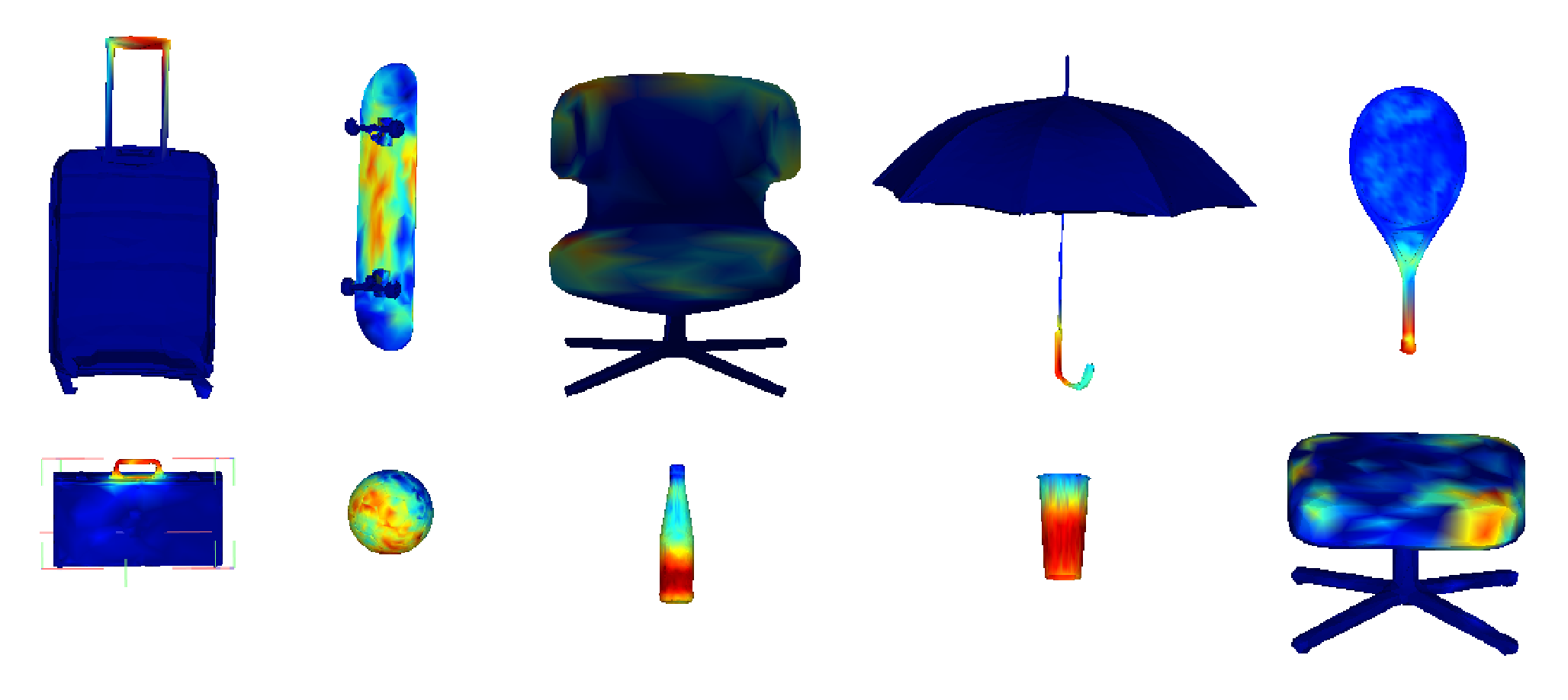}  \vrule
    \includegraphics[height=0.236 \linewidth,trim={0mm 10mm 10mm 14mm}, clip=true]{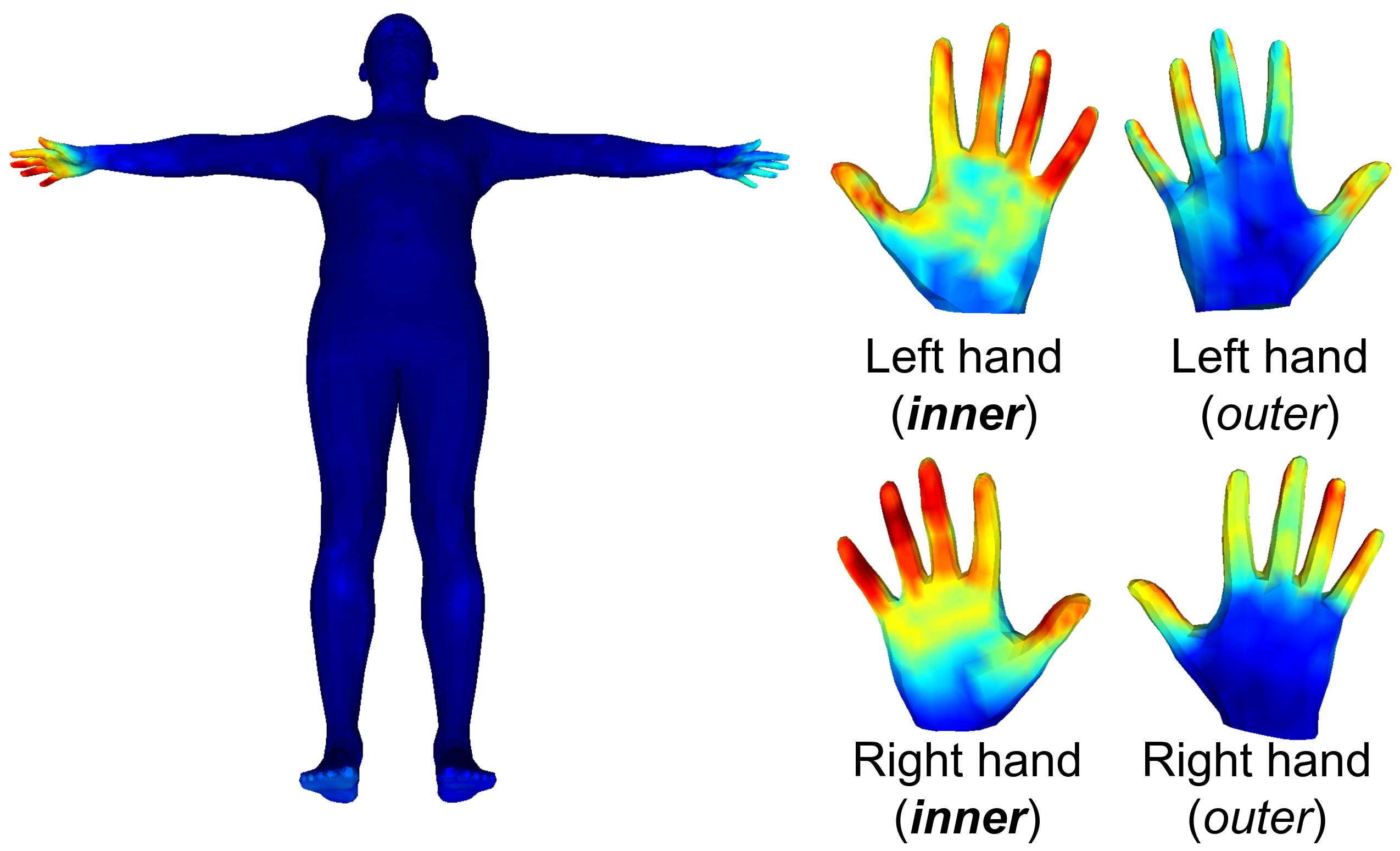}
    \caption{
                    Contact heatmaps for \CRColor{each object}
                    (across all subjects) 
                    and the human body (across all objects and subjects). 
                    \CRColor{Contact likelihood is color-coded;} high likelihood is shown with red, and low with blue. 
                    Color-coding is normalized separately for each object, the body, and each hand. 
    }
    \label{fig:contact_heatmaps}
\end{figure}

\section{Experiments}   \label{sec:experiments}

\textbf{Contact Heatmaps.}
\Fig{fig:contact_heatmaps}\ourrefcolor{-left} 
shows contact heatmaps on each object, across all subjects. 
We follow the protocol of \grab~\cite{taheri2020grab}, which uses a proximity metric on reconstructed human and object meshes. 
First, we compute per-frame binary contact maps by thresholding \REBT{(at 4.5mm)}
the distances from each body vertex to the closest object surface point. 
Then, we integrate these maps over time (and subjects) to get ``heatmaps'' encoding contact likelihood. 
\ourname reconstructs human and object meshes accurately enough so that contact heatmaps agree with 
object affordances, \eg, the handle of the suitcase, umbrella and tennis racquet are likely to be grasped, the upper skateboard surface is likely to be contacted by the foot, and the upper stool surface 
by the buttocks.

\Fig{fig:contact_heatmaps}\ourrefcolor{-right} shows heatmaps on the body, computed across all subjects and objects. Heatmaps show that most \CRColor{of \ourname's} interactions involve mainly the right hand. Contact on the palm looks realistic, and is concentrated on the fingers and MCP joints. \CRColor{The ``false''} contact on the dorsal side is attributed to our challenging \CRColor{camera} setup \CRColor{and interaction scenarios, as well as} some reconstruction jitter.

\begin{figure}[t]
    \centering
    \includegraphics[width=\linewidth, trim={00mm 00mm 14mm 10mm}, clip]{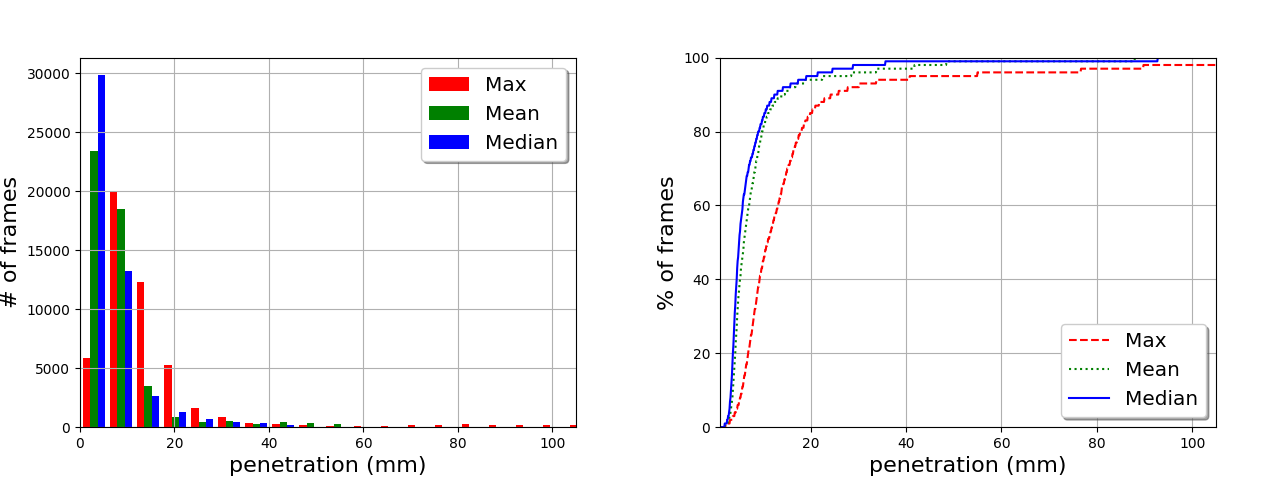}
    \caption{
                Statistics of human-object mesh penetration 
                for all \ourname sequences. \textbf{Left:} The number of frames (Y-axis) with a certain penetration depth (X-axis). \textbf{Right:} The percentage of frames (Y-axis) with a penetration depth below a threshold (X-axis). \CRColor{In the legend, ``Max'', ``Mean'' and ``Median'' refer to three ways of reporting the penetration for each frame, \ie, taking the maximum, mean and median value of the penetration depth of all vertices, respectively.}
    }
    \label{fig:penetration_plots}
\end{figure}

\textbf{Penetration.}
We evaluate the penetration between human and object mesh\-es for all sequences of our dataset.
\REBT{We follow the protocol of \grab~\etal~\cite{taheri2020grab}; we first find the ``contact frames'' for which there is at least minimal human-object contact, and then report statistics for these.} In \fig{fig:penetration_plots}\ourrefcolor{-left} we show the distribution of  penetrations, \ie, the number of ``contact frames'' \CRColor{(Y axis)} with a \CRColor{certain} 
mesh penetration depth \CRColor{(X axis)}. In \fig{fig:penetration_plots}\ourrefcolor{-right} we show the cumulative distribution of penetration, \ie, the percentage of ``contact frames'' \CRColor{(Y axis)} for which mesh penetration is below a threshold \CRColor{(X axis)}. Roughly $60\%$ of ``contact frames'' have $\leq5mm$, $80\%$ $\leq7$ mm, and  $98\%$ $\leq20$ mm \CRColor{mean} penetration. The average penetration depth over all ``contact frames'' is $7.2$ mm.

\textbf{Fitting Accuracy.}
For every frame, we compute the distance from each mesh vertex to the closest point-cloud (PCL) point; for each human or object mesh we take into account only the respective PCL area obtained with \pointrend~\cite{kirillov2020pointrend} segmentation. The mean vertex-to-PCL distance is 20.29 mm for the body, and 18.50 mm for objects. \REBT{In comparison, \proxD \cite{hassan2019resolving}, our base method, achieves an error of 13.02 mm for the body. This is expected since \proxD is free to change the body shape to fit each individual frame, while our method estimates a single body shape for the whole sequence. SMPLify-X \cite{pavlakos2019expressive} achieves an mean error of 79.54 mm, for VIBE the mean error is 55.59 mm, while ExPose gets an mean error of 71.78 mm. These numbers validate the effectiveness of our method for body tracking. Note that these methods are based on monocular \rgb images only, so there is not enough information for them to accurately estimate the global position of the \threeD body meshes. Thus we first align the output meshes with the point clouds, then compute the error.} \REBT{Note that the error is bounded from below for two reasons: 
(1) it is  influenced by factory-design imperfections in the synchronization of \kinectAZs}, and 
(2) some vertices reflect body/object areas that are occluded during interaction and their closest PCL point is a wrong correspondence. \REBT{Despite this, \ourname empirically estimates reasonable bodies, hands and objects in interaction,} \REBT{as reflected in the contact heatmaps and penetration metrics discussed above.}

\begin{figure}[t]
    \centering
    \includegraphics[trim=014mm 022mm 035mm 018mm, clip=true, height=0.274 \textwidth]{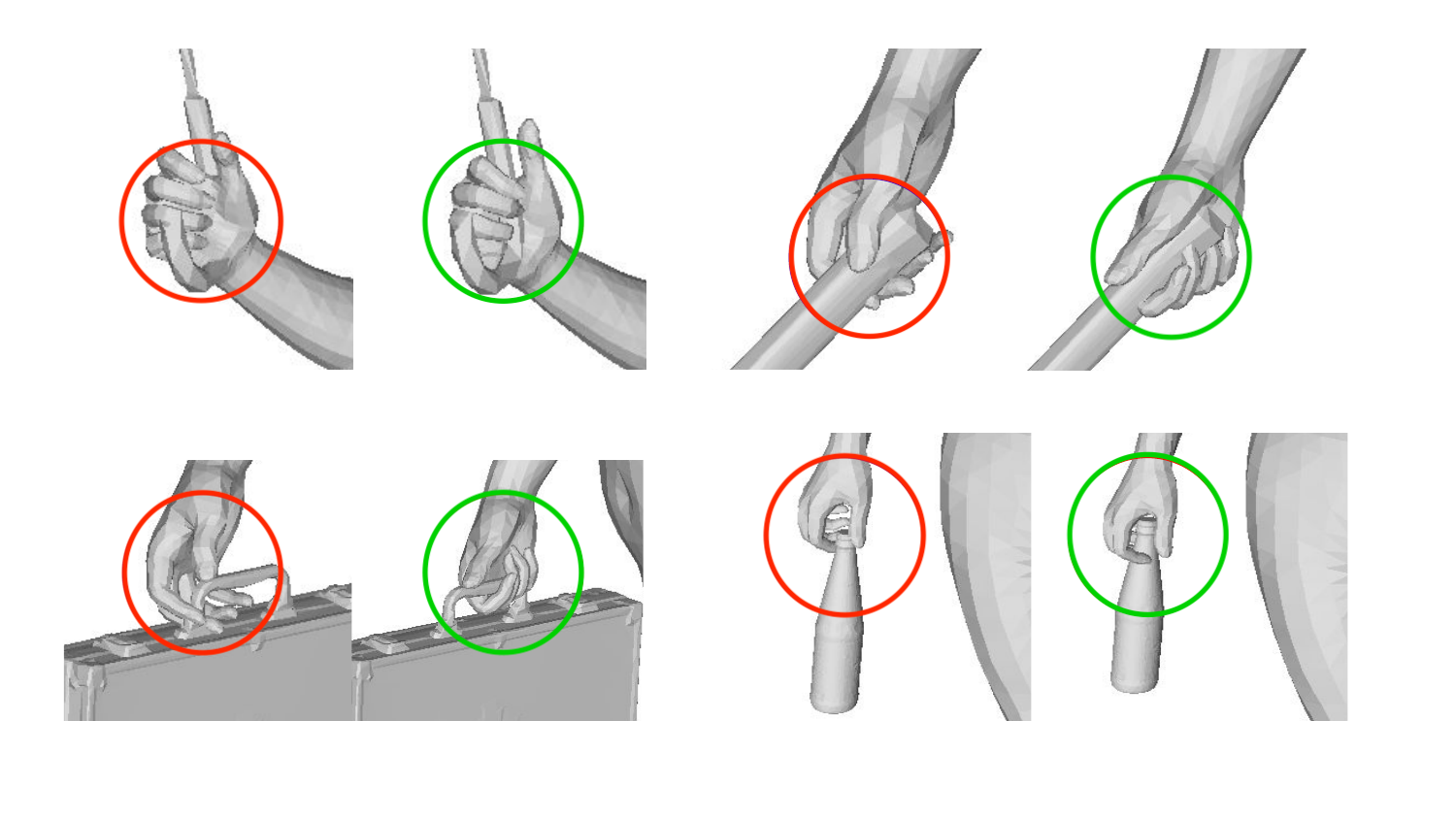}
    \quad
    \includegraphics[trim=000mm 008mm 014mm 012mm, clip=true, height=0.284 \textwidth]{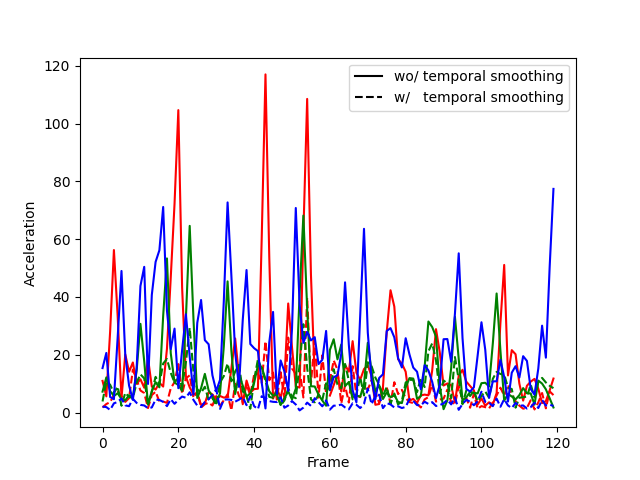}
    \caption{
    \REBT{
    \textbf{Left:} 
    Qualitative ablation of our contact term. 
    Each pair of images shows results wo/ (red) and w/ (green) the contact term. 
    Encouraging
    contact results in more natural hand poses and hand-object grasps. 
    \textbf{Right:} 
    Acceleration 
    of a random vertex w/ (dashed line) and wo/ (solid line) temporal smoothing for 3 sequences \CRColor{(shown with different color)} over the first 120 frames. 
    \CRColor{Dashed lines (w/ temporal smoothing) correspond to lower acceleration, \ie, less jitter.} 
    }
    }
    \label{fig:contact}
\end{figure}

\REBT{
\textbf{Ablation of Contact Term.} \Fig{fig:contact}\ourrefcolor{-left} shows results with-/out our term that encourages body-object contact; \CRColor{visualization``zooms'' into} hand-object grasps. We see that encouraging contact yields more natural hand poses and fewer interpenetrations. This is backed up by the contact heatmaps and penetration metrics discussed above.
}

\REBT{ \textbf{Ablation of Temporal Smoothing Term.} \Fig{fig:contact}\ourrefcolor{-right} shows results with-/out our temporal smoothing term. Each solid line shows the acceleration of a randomly chosen vertex without the temporal smoothness term; we show 3 different motions. The dashed lines of the same color show the same motions with the smoothness term; these are clearly smoother. } \REBT{We \CRColor{empirically} find that the learned motion prior of Zhang \etal~\cite{zhang2021learning} produces a more natural motion than handcrafted ones \cite{huang2017towards}. 
}

\CRColor{\textbf{Discussion on Jitter.}} \CRColor{Despite the smoothing, some jitter is still inevitable. We attribute this to two factors: (1) \openpose and Mask-RCNN are empirically relatively sensitive to occlusions and illumination (\eg, reflections, shadows, poor lighting); the data terms for fitting \threeD models depend on these. (2) \kinectAZs have a reasonable synchronization, yet, there is still a small delay among cameras to avoid depth-camera interference; the point cloud ``gathered'' across views is a bit ``patchy'' as information pieces have a small time difference.} The jitter is more intense for hands relatively to the body, due to their low image resolution, motion blur, and coarse point clouds. \CRColor{Despite these challenges, \ourname is a good step towards capturing everyday whole-body interactions with commodity hardware. } Future work will study advanced motion priors.
     
\section{Discussion}    \label{sec:conclusion}

Here we focus on \CRColor{whole}-body human interaction with everyday rigid objects. We present a novel method, called \ourname, that reconstructs such interactions from multi-view full-body videos, including natural hand poses and contact with objects. With this method, we capture the \CRColor{novel} \ourname dataset, \CRColor{with} a variety of people interacting with several common objects. The dataset contains reconstructed \threeD meshes for the \CRColor{whole body} and the object over time \CRColor{(\ie, \fourD meshes)}, as well as plausible contacts between them. In contrast to most previous work, our method uses no special devices like optical markers or IMUs, but only several consumer-level \rgbD cameras. Our setup is lightweight and has the potential \CRColor{to be used} in daily scenarios. Our method estimates reasonable hand poses even when there is heavy occlusion between hands and the object. \CRColor{In future work, we plan to study interactions with smaller objects and dexterous manipulation.} Our data and code are available at \websiteURL. 

\section{Acknowledgements}
We thank Chun-Hao P. Huang, Hongwei Yi, Jiaxiang Shang, and Mohamed Hassan for helpful discussion of technical details. We thank Yuliang Xiu, Jinlong Yang, Victoria F. Abrevaya, Taylor McConnell, Galina Henz, Marku H{\"o}schle, Senya Polikovsky, Matvey Safroshkin and Tsvetelina Alexiadis for data collection and cleaning. We thank all the participants of our experiments, and Benjamin Pellkofer for IT and website support. 

\noindent
\CRColor{The authors thank the International Max Planck Research School for Intelligent Systems (IMPRS-IS) for supporting OT. This work was supported by the German Federal Ministry of Education and Research (BMBF): T{\"u}bingen AI Center, FKZ: 01IS18039B.}

\noindent
\CRColor{\textbf{Conflict of Interest.}}
Disclosure: \url{https://files.is.tue.mpg.de/black/CoI_GCPR_2022.txt}.

{\noindent \large \bf {APPENDIX}}\\

\appendix
\setcounter{equation}{0}
\renewcommand{\thefigure}{A.\arabic{figure}}
\renewcommand{\thetable}{A.\arabic{table}}
\renewcommand{\theequation}{A.\arabic{equation}}

\section{Video \CRColorSUPMAT{on our Website}}
The \CRColorSUPMAT{narrated} 
video \CRColorSUPMAT{on our website} (\websiteURL) presents:
\begin{itemize}
    \item 
    \CRColorSUPMAT{An explanation of our motivation.}
    \item 
    \CRColorSUPMAT{An overview of our \ourname dataset and method.}
    \item 
    Some \CRColorSUPMAT{videos (input) and reconstructed \fourD meshes} 
    \CRColorSUPMAT{(output)} of our \ourname dataset.
    \item A \CRColorSUPMAT{qualitative} comparison between our \ourname mesh reconstructions to the ones from \smplifyX~\cite{pavlakos2019expressive}, \expose~\cite{choutas2020monocular}, and \vibe~\cite{kocabas2020vibe}.
\end{itemize}

\section{Optimization \CRColorSUPMAT{Objective Function \& Terms}}
We use the \CRColorSUPMAT{objective} function of Eq.~\textcolor{red}{6} of the main paper to jointly refine (via optimization) the body and object motion over the whole sequence. \CRColorSUPMAT{Here} we give a detailed explanation of the terms not elaborated in the main \CRColorSUPMAT{paper} due to space limitations.

The \CRColorSUPMAT{motion smoothness} term $E_{\mathcal{S}}$ penalizes sudden changes in the \CRColorSUPMAT{position of body vertices}. \CRColorSUPMAT{It employs the learned motion prior of LEMO~\cite{zhang2021learning}} and
is defined as: 
\begin{align}
\begin{split}
\CRColorSUPMAT{
    E_{\mathcal{S}}(\Theta, \Psi, \Gamma, A; T, \beta^*) = 
    \frac{1}{Q(T-2)} \sum_{t=1}^{T-1} 
    \big \| 
    z_{t+1}^{opt} - z_t^{opt}
    \big \|^2,
}
\end{split}
\end{align}
\CRColorSUPMAT{where $T$ is the sequence length, $Q$ is a constant representing the number of virtual body-markers of LEMO (see \cite{zhang2021learning} for an explanation; they use a different symbol), $z_t^{opt}$ is the latent vector for the $t$-th frame from LEMO's pre-trained motion auto-encoder ($F_S$): 
\begin{equation}
     Z^{opt} = F_S(X_{\Delta}^{opt}) = [z_1^{opt}, z_2^{opt}, ..., z_{T-1}^{opt}],
\end{equation}
where $X_{\Delta}^{opt}$ is a (concatenated) vector containing the temporal position change of LEMO's virtual body-markers. 
For more details, please refer to the paper of LEMO~\cite{zhang2021learning}.}

The \CRColorSUPMAT{vertex acceleration} term $E_{\mathcal{A}}$ \CRColorSUPMAT{is a simple hand-crafted motion prior that encourages smooth motion trajectories for the object:} 
\begin{align}
\begin{split}
    E_{\mathcal{A}}(\Xi; T, M) = 
    \frac{1}{T-2} \sum_{t=2}^{T-1} 
    \Big \| 
    &
    W'(\Xi_{t-1}, M) + 
    \\ & 
    W'(\Xi_{t+1}, M) - 
    2 * W'(\Xi_t, M) 
    \Big \|^2
\end{split}
\end{align}
where $M$ is the \CRColorSUPMAT{object mesh}, and $W'$ \CRColorSUPMAT{denotes} the operation of \CRColorSUPMAT{first} rigidly deforming the object according to $\Xi_t$ \CRColorSUPMAT{and then} concatenating the vertices into a single vector. 

The \CRColorSUPMAT{contact} term $E_{\mathcal{C}}(\beta^*, \Theta_t, \Psi_t, \Gamma_t, \Xi_t, M)$ \CRColorSUPMAT{encourages} the \CRColorSUPMAT{annotated likely contact areas of the body (see Fig.~\textcolor{red}{3} of the main paper)} to be in contact with the 
\CRColorSUPMAT{object:}
\begin{align}
\begin{split}
    E_{\mathcal{C}}(\beta^*, \Theta_t, \Psi_t, \Gamma_t, \Xi_t, M) = 
        CD 
        \Big ( 
        & H  \big(   
                    W(\Theta_{t}, \Psi_{t}, \Gamma_{t}, A, \beta^*)
             \big), \\
        & H' \big(   
                    W'(\Xi_{t}, M)
             \big)
        \Big )
\end{split}
\end{align}
where 
$\text{CD}$ refers to the Chamfer Distance function, 
\CRColorSUPMAT{$H$ is a function that returns only the annotated body-contact vertices of Fig.~\textcolor{red}{3},  $H'$ returns the closest points on the object for these body-contact vertices, $W'$ deforms rigidly the object and is explained in the previous paragraph and $W$ similarly (non-rigidly) deforms the \smplX mesh and concatenates the vertices into a single vector.
}

\CRColorSUPMAT{Finally, the ground-support terms $E_\mathcal{G}$ and $E_\mathcal{G'}$ build on} the fact that \CRColorSUPMAT{no} human or object vertex, \CRColorSUPMAT{respectively,} should be below the ground plane, \CRColorSUPMAT{and penalize any vertex penetrating the ground}. 
Let $p_\mathcal{G}$ be a point on the \CRColorSUPMAT{ground} plane and $n_{\mathcal{G}}$ be the \CRColorSUPMAT{corresponding} normal; \CRColorSUPMAT{both defined are once and offline}. 
Then the term $E_\mathcal{G}$ for body-ground penetration is defined as:
\begin{equation}
    E_{\mathcal{G}}(\beta^*, \Theta_t, \Psi_t, \Gamma_t) = 
    \Big \|
    RL  
    \Big ( 
            n_{\mathcal{G}} * 
            \big ( 
                    p_{\mathcal{G}} - W(\beta^*, \Theta_t, \Psi_t, \Gamma_t)
            \big )
    \Big ) 
    \Big \|^2,
\end{equation}
where $RL$ is the ReLU function, and $*$ here is the inner  product of vectors. 
The term $E_{\mathcal{G}'}$ for object-ground penetration follows a similar formulation:
\begin{equation}
    E_{\mathcal{G'}}(\Xi_t, M) = 
    \Big \| 
            RL( n_{\mathcal{G}} * ( p_{\mathcal{G}} - W'(\Xi_{t}, M) )  ) 
    \Big \|^2.
\end{equation}


\bibliographystyle{splncs04}
\bibliography{MAIN_PAPER}

\begin{thebibliography}{10}
\providecommand{\url}[1]{\texttt{#1}}
\providecommand{\urlprefix}{URL }
\providecommand{\doi}[1]{https://doi.org/#1}

\bibitem{alldieck2018video}
Alldieck, T., Magnor, M., Xu, W., Theobalt, C., Pons-Moll, G.: Video based
  reconstruction of {3D} people models. In: {Computer Vision and Pattern
  Recognition (CVPR)}. pp. 8387--8397 (2018)

\bibitem{anguelov2005scape}
Anguelov, D., Srinivasan, P., Koller, D., Thrun, S., Rodgers, J., Davis, J.:
  {SCAPE}: {S}hape completion and animation of people. {Transactions on
  Graphics (TOG)}  \textbf{24}(3),  408--416 (2005)

\bibitem{bhatnagar2022behave}
Bhatnagar, B.L., Xie, X., Petrov, I.A., Sminchisescu, C., Theobalt, C.,
  Pons-Moll, G.: {BEHAVE}: {D}ataset and method for tracking human object
  interactions. In: {Computer Vision and Pattern Recognition (CVPR)}. pp.
  15935--15946 (2022)

\bibitem{bogo2016keep}
Bogo, F., Kanazawa, A., Lassner, C., Gehler, P., Romero, J., Black, M.J.: Keep
  it {SMPL}: {A}utomatic estimation of {3D} human pose and shape from a single
  image. In: {European Conference on Computer Vision (ECCV)}. vol.~9909, pp.
  561--578 (2016)

\bibitem{cao2020long}
Cao, Z., Gao, H., Mangalam, K., Cai, Q., Vo, M., Malik, J.: Long-term human
  motion prediction with scene context. In: {European Conference on Computer
  Vision (ECCV)}. vol. 12346, pp. 387--404 (2020)

\bibitem{cao2019openpose}
Cao, Z., Hidalgo, G., Simon, T., Wei, S.E., Sheikh, Y.: {OpenPose}: {R}ealtime
  multi-person {2D} pose estimation using part affinity fields. {Transactions
  on Pattern Analysis and Machine Intelligence (TPAMI)}  \textbf{43}(1),
  172--186 (2019)

\bibitem{choutas2020monocular}
Choutas, V., Pavlakos, G., Bolkart, T., Tzionas, D., Black, M.J.: Monocular
  expressive body regression through body-driven attention. In: {European
  Conference on Computer Vision (ECCV)}. vol. 12355, pp. 20--40 (2020)

\bibitem{de2008performance}
De~Aguiar, E., Stoll, C., Theobalt, C., Ahmed, N., Seidel, H.P., Thrun, S.:
  Performance capture from sparse multi-view video. {Transactions on Graphics
  (TOG)}  \textbf{27}(3),  1--10 (2008)

\bibitem{Dong_2021_PAMI}
Dong, J., Fang, Q., Jiang, W., Yang, Y., Huang, Q., Bao, H., Zhou, X.: Fast and
  robust multi-person {3D} pose estimation and tracking from multiple views.
  {Transactions on Pattern Analysis and Machine Intelligence (TPAMI)}
  \textbf{14}(8),  1--12 (2021)

\bibitem{Dong_2019_CVPR}
Dong, J., Jiang, W., Huang, Q., Bao, H., Zhou, X.: Fast and robust multi-person
  {3D} pose estimation from multiple views. In: {Computer Vision and Pattern
  Recognition (CVPR)}. pp. 7792--7801 (2019)

\bibitem{dong2021shape}
Dong, Z., Song, J., Chen, X., Guo, C., Hilliges, O.: Shape-aware multi-person
  pose estimation from multi-view images. In: {International Conference on
  Computer Vision ({ICCV})}. pp. 11158--11168 (2021)

\bibitem{GemanMcClure1987}
Geman, S., McClure, D.E.: Statistical methods for tomographic image
  reconstruction. In: Proceedings of the 46th Session of the International
  Statistical Institute, Bulletin of the ISI. vol.~52 (1987)

\bibitem{Hamer_Hand_Manipulating_2009}
Hamer, H., Schindler, K., Koller-Meier, E., {Van Gool}, L.: Tracking a hand
  manipulating an object. In: {International Conference on Computer Vision
  ({ICCV})}. pp. 1475--1482 (2009)

\bibitem{hampali2020honnotate}
Hampali, S., Rad, M., Oberweger, M., Lepetit, V.: {HOnnotate}: {A} method for
  {3D} annotation of hand and object poses. In: {Computer Vision and Pattern
  Recognition (CVPR)}. pp. 3193--3203 (2020)

\bibitem{hassan2019resolving}
Hassan, M., Choutas, V., Tzionas, D., Black, M.J.: Resolving {3D} human pose
  ambiguities with {3D} scene constrains. In: {International Conference on
  Computer Vision ({ICCV})}. pp. 2282--2292 (2019)

\bibitem{Hassan2021posa}
Hassan, M., Ghosh, P., Tesch, J., Tzionas, D., Black, M.J.: Populating {3D}
  scenes by learning human-scene interaction. In: {Computer Vision and Pattern
  Recognition (CVPR)}. pp. 14708--14718 (2021)

\bibitem{hasson2020leveraging}
Hasson, Y., Tekin, B., Bogo, F., Laptev, I., Pollefeys, M., Schmid, C.:
  Leveraging photometric consistency over time for sparsely supervised
  hand-object reconstruction. In: {Computer Vision and Pattern Recognition
  (CVPR)}. pp. 568--577 (2020)

\bibitem{hasson2019learning}
Hasson, Y., Varol, G., Tzionas, D., Kalevatykh, I., Black, M.J., Laptev, I.,
  Schmid, C.: Learning joint reconstruction of hands and manipulated objects.
  In: {Computer Vision and Pattern Recognition (CVPR)}. pp. 11807--11816 (2019)

\bibitem{epipolartransformers2020cvpr}
He, Y., Yan, R., Fragkiadaki, K., Yu, S.I.: Epipolar transformers. In:
  {Computer Vision and Pattern Recognition (CVPR)}. pp. 7776--7785 (2020)

\bibitem{hu2019sail}
Hu, Y.T., Chen, H.S., Hui, K., Huang, J.B., Schwing, A.G.: {SAIL-VOS}:
  {S}emantic amodal instance level video object segmentation - a synthetic
  dataset and baselines. In: {Computer Vision and Pattern Recognition (CVPR)}.
  pp. 3105--3115 (2019)

\bibitem{Huang:CVPR:2022}
Huang, C.H.P., Yi, H., H{\"o}schle, M., Safroshkin, M., Alexiadis, T.,
  Polikovsky, S., Scharstein, D., Black, M.J.: Capturing and inferring dense
  full-body human-scene contact. In: {Computer Vision and Pattern Recognition
  (CVPR)}. pp. 13274--13285 (2022)

\bibitem{huang2017towards}
Huang, Y., Bogo, F., Lassner, C., Kanazawa, A., Gehler, P.V., Romero, J.,
  Akhter, I., Black, M.J.: Towards accurate marker-less human shape and pose
  estimation over time. In: {International Conference on 3D Vision (3DV)}. pp.
  421--430 (2017)

\bibitem{ionescu2013human3}
Ionescu, C., Papava, D., Olaru, V., Sminchisescu, C.: Human3.6{M}: {L}arge
  scale datasets and predictive methods for {3D} human sensing in natural
  environments. {Transactions on Pattern Analysis and Machine Intelligence
  (TPAMI)}  \textbf{36}(7),  1325--1339 (2014)

\bibitem{iskakov2019learnable}
Iskakov, K., Burkov, E., Lempitsky, V., Malkov, Y.: Learnable triangulation of
  human pose. In: {International Conference on Computer Vision ({ICCV})}. pp.
  7717--7726 (2019)

\bibitem{kanazawa2018end}
Kanazawa, A., Black, M.J., Jacobs, D.W., Malik, J.: End-to-end recovery of
  human shape and pose. In: {Computer Vision and Pattern Recognition (CVPR)}.
  pp. 7122--7131 (2018)

\bibitem{karunratanakul2020grasping}
Karunratanakul, K., Yang, J., Zhang, Y., Black, M.J., Muandet, K., Tang, S.:
  {Grasping Field}: {L}earning implicit representations for human grasps. In:
  {International Conference on 3D Vision (3DV)}. pp. 333--344 (2020)

\bibitem{kato2018neural}
Kato, H., Ushiku, Y., Harada, T.: Neural {3D} mesh renderer. In: {Computer
  Vision and Pattern Recognition (CVPR)}. pp. 3907--3916 (2018)

\bibitem{kirillov2020pointrend}
Kirillov, A., Wu, Y., He, K., Girshick, R.: {PointRend}: {I}mage segmentation
  as rendering. In: {Computer Vision and Pattern Recognition (CVPR)}. pp.
  9799--9808 (2020)

\bibitem{kocabas2020vibe}
Kocabas, M., Athanasiou, N., Black, M.J.: {VIBE}: {V}ideo inference for human
  body pose and shape estimation. In: {Computer Vision and Pattern Recognition
  (CVPR)}. pp. 5252--5262 (2020)

\bibitem{li2019putting}
Li, X., Liu, S., Kim, K., Wang, X., Yang, M., Kautz, J.: Putting humans in a
  scene: {L}earning affordance in {3D} indoor environments. In: {Computer
  Vision and Pattern Recognition (CVPR)}. pp. 12368--12376 (2019)

\bibitem{liu2011markerless}
Liu, Y., Stoll, C., Gall, J., Seidel, H.P., Theobalt, C.: Markerless motion
  capture of interacting characters using multi-view image segmentation. In:
  {Computer Vision and Pattern Recognition (CVPR)}. pp. 1249--1256 (2011)

\bibitem{loper2014mosh}
Loper, M., Mahmood, N., Black, M.J.: {MoSh}: {M}otion and shape capture from
  sparse markers. {Transactions on Graphics (TOG)}  \textbf{33}(6),  1--13
  (2014)

\bibitem{loper2015smpl}
Loper, M., Mahmood, N., Romero, J., Pons-Moll, G., Black, M.J.: {SMPL}: {A}
  skinned multi-person linear model. {Transactions on Graphics (TOG)}
  \textbf{34}(6),  248:1--248:16 (2015)

\bibitem{loper2014opendr}
Loper, M.M., Black, M.J.: {OpenDR}: {A}n approximate differentiable renderer.
  In: {European Conference on Computer Vision (ECCV)}. vol.~8695, pp. 154--169
  (2014)

\bibitem{mahmood2019amass}
Mahmood, N., Ghorbani, N., F.~Troje, N., Pons-Moll, G., Black, M.J.: {AMASS}:
  {A}rchive of motion capture as surface shapes. In: {International Conference
  on Computer Vision ({ICCV})}. pp. 5441--5450 (2019)

\bibitem{von2018recovering}
von Marcard, T., Henschel, R., Black, Michael J.and~Rosenhahn, B., Pons-Moll,
  G.: Recovering accurate {3D} human pose in the wild using {IMUs} and a moving
  camera. In: {European Conference on Computer Vision (ECCV)}. vol. 11214, pp.
  614--631 (2018)

\bibitem{mehta2017vnect}
Mehta, D., Sridhar, S., Sotnychenko, O., Rhodin, H., Shafiei, M., Seidel, H.P.,
  Xu, W., Casas, D., Theobalt, C.: {VNect}: Real-time {3D} human pose
  estimation with a single {RGB} camera. {Transactions on Graphics (TOG)}
  \textbf{36}(4),  44:1--44:14 (2017)

\bibitem{kinectAPI}
Microsoft: {Azure Kinect SDK (K4A)}.
  https://github.com/microsoft/Azure-Kinect-Sensor-SDK (2022)

\bibitem{newell2016stacked}
Newell, A., Yang, K., Deng, J.: Stacked hourglass networks for human pose
  estimation. In: {European Conference on Computer Vision (ECCV)}. vol.~9912,
  pp. 483--499 (2016)

\bibitem{oikonomidis2011full}
Oikonomidis, I., Kyriazis, N., Argyros, A.A.: Full {DOF} tracking of a hand
  interacting with an object by modeling occlusions and physical constraints.
  In: {International Conference on Computer Vision ({ICCV})}. pp. 2088--2095
  (2011)

\bibitem{omran2018neural}
Omran, M., Lassner, C., Pons-Moll, G., Gehler, P., Schiele, B.: Neural body
  fitting: {U}nifying deep learning and model based human pose and shape
  estimation. In: {International Conference on 3D Vision (3DV)}. pp. 484--494
  (2018)

\bibitem{supr2022}
Osman, A.A.A., Bolkart, T., Tzionas, D., Black, M.J.: {SUPR}: {A} sparse
  unified part-based human body model. In: {European Conference on Computer
  Vision (ECCV)} (2022)

\bibitem{osman2020star}
Osman, A.A., Bolkart, T., Black, M.J.: {STAR}: {S}parse trained articulated
  human body regressor. In: {European Conference on Computer Vision (ECCV)}.
  vol. 12351, pp. 598--613 (2020)

\bibitem{pavlakos2019expressive}
Pavlakos, G., Choutas, V., Ghorbani, N., Bolkart, T., Osman, A.A.A., Tzionas,
  D., Black, M.J.: Expressive body capture: {3D} hands, face, and body from a
  single image. In: {Computer Vision and Pattern Recognition (CVPR)}. pp.
  10975--10985 (2019)

\bibitem{pons2010multisensor}
Pons-Moll, G., Baak, A., Helten, T., M{\"u}ller, M., Seidel, H.P., Rosenhahn,
  B.: Multisensor-fusion for {3D} full-body human motion capture. In: {Computer
  Vision and Pattern Recognition (CVPR)}. pp. 663--670 (2010)

\bibitem{Qiu_2019_ICCV}
Qiu, H., Wang, C., Wang, J., Wang, N., Zeng, W.: Cross view fusion for {3D}
  human pose estimation. In: {International Conference on Computer Vision
  ({ICCV})}. pp. 4341--4350 (2019)

\bibitem{rhodin2016general}
Rhodin, H., Robertini, N., Casas, D., Richardt, C., Seidel, H.P., Theobalt, C.:
  General automatic human shape and motion capture using volumetric contour
  cues. In: {European Conference on Computer Vision (ECCV)}. vol.~9909, pp.
  509--526 (2016)

\bibitem{Rogez2015everyday}
Rogez, G., III, J.S.S., Ramanan, D.: Understanding everyday hands in action
  from {RGB-D} images. In: {International Conference on Computer Vision
  ({ICCV})}. pp. 3889--3897 (2015)

\bibitem{romero2010handsInAction}
Romero, J., Kjellstr{\"{o}}m, H., Kragic, D.: Hands in action: {R}eal-time {3D}
  reconstruction of hands in interaction with objects. In: {International
  Conference on Robotics and Automation (ICRA)}. pp. 458--463 (2010)

\bibitem{romero2017embodied}
Romero, J., Tzionas, D., Black, M.J.: Embodied hands: {M}odeling and capturing
  hands and bodies together. {Transactions on Graphics (TOG)}  \textbf{36}(6),
  245:1--245:17 (2017)

\bibitem{savva2016pigraphs}
Savva, M., Chang, A.X., Hanrahan, P., Fisher, M., Nie{\ss}ner, M.: {PiGraphs}:
  {L}earning interaction snapshots from observations. {Transactions on Graphics
  (TOG)}  \textbf{35}(4),  139:1--139:12 (2016)

\bibitem{sigal2006humaneva}
Sigal, L., Balan, A., Black, M.J.: {HumanEva}: {S}ynchronized video and motion
  capture dataset and baseline algorithm for evaluation of articulated human
  motion. {International Journal of Computer Vision (IJCV)}  \textbf{87}(1-2),
  4--27 (2010)

\bibitem{sun2022onepose}
Sun, J., Wang, Z., Zhang, S., He, X., Zhao, H., Zhang, G., Zhou, X.: {OnePose}:
  {O}ne-shot object pose estimation without {CAD} models. In: CVPR. pp.
  6825--6834 (2022)

\bibitem{taheri2020grab}
Taheri, O., Ghorbani, N., Black, M.J., Tzionas, D.: {GRAB}: {A} dataset of
  whole-body human grasping of objects. In: {European Conference on Computer
  Vision (ECCV)}. vol. 12349, pp. 581--600 (2020)

\bibitem{to2020voxelpose}
Tu, H., Wang, C., Zeng, W.: {VoxelPose}: Towards multi-camera {3D} human pose
  estimation in wild environment. In: {European Conference on Computer Vision
  (ECCV)}. vol. 12346, pp. 197--212 (2020)

\bibitem{tzionas2016capturing}
Tzionas, D., Ballan, L., Srikantha, A., Aponte, P., Pollefeys, M., Gall, J.:
  Capturing hands in action using discriminative salient points and physics
  simulation. {International Journal of Computer Vision (IJCV)}
  \textbf{118}(2),  172--193 (2016)

\bibitem{varol2017long}
Varol, G., Laptev, I., Schmid, C.: Long-term temporal convolutions for action
  recognition. {Transactions on Pattern Analysis and Machine Intelligence
  (TPAMI)}  \textbf{40}(6),  1510--1517 (2017)

\bibitem{wei2016convolutional}
Wei, S.E., Ramakrishna, V., Kanade, T., Sheikh, Y.: Convolutional pose
  machines. In: {Computer Vision and Pattern Recognition (CVPR)}. pp.
  4724--4732 (2016)

\bibitem{xu2020ghum}
Xu, H., Bazavan, E.G., Zanfir, A., Freeman, W.T., Sukthankar, R., Sminchisescu,
  C.: {GHUM} {\&} {GHUML}: Generative {3D} human shape and articulated pose
  models. In: {Computer Vision and Pattern Recognition (CVPR)}. pp. 6183--6192
  (2020)

\bibitem{xu2018monoperfcap}
Xu, W., Chatterjee, A., Zollh{\"o}fer, M., Rhodin, H., Mehta, D., Seidel, H.P.,
  Theobalt, C.: {MonoPerfCap}: {H}uman performance capture from monocular
  video. {Transactions on Graphics (TOG)}  \textbf{37}(2),  1--15 (2018)

\bibitem{yao2010modeling}
Yao, B., Fei-Fei, L.: Modeling mutual context of object and human pose in
  human-object interaction activities. In: {Computer Vision and Pattern
  Recognition (CVPR)}. pp. 17--24 (2010)

\bibitem{yi2022mover}
Yi, H., Huang, C.H.P., Tzionas, D., Kocabas, M., Hassan, M., Tang, S., Thies,
  J., Black, M.J.: Human-aware object placement for visual environment
  reconstruction. In: {Computer Vision and Pattern Recognition (CVPR)}. pp.
  3959--3970 (2022)

\bibitem{zhang2020phosa}
Zhang, J.Y., Pepose, S., Joo, H., Ramanan, D., Malik, J., Kanazawa, A.:
  Perceiving {3D} human-object spatial arrangements from a single image in the
  wild. In: {European Conference on Computer Vision (ECCV)} (2020)

\bibitem{zhang2021learning}
Zhang, S., Zhang, Y., Bogo, F., Pollefeys, M., Tang, S.: Learning motion priors
  for {4D} human body capture in {3D} scenes. In: {Computer Vision and Pattern
  Recognition (CVPR)}. pp. 11323--11333 (2021)

\bibitem{zhang2020generating}
Zhang, Y., Hassan, M., Neumann, H., Black, M.J., Tang, S.: Generating {3D}
  people in scenes without people. In: {Computer Vision and Pattern Recognition
  (CVPR)}. pp. 6193--6203 (2020)

\bibitem{Zhang_2020_CVPR}
Zhang, Y., An, L., Yu, T., Li, X., Li, K., Liu, Y.: {4D} association graph for
  realtime multi-person motion capture using multiple video cameras. In:
  {Computer Vision and Pattern Recognition (CVPR)}. pp. 1321--1330 (2020)

\bibitem{lightcap2021}
Zhang, Y., Li, Z., An, L., Li, M., Yu, T., Liu, Y.: Light-weight multi-person
  total capture using sparse multi-view cameras. In: {International Conference
  on Computer Vision ({ICCV})}. pp. 5560--5569 (2021)

\bibitem{zollhofer2018state}
Zollh{\"o}fer, M., Stotko, P., G{\"o}rlitz, A., Theobalt, C., Nie{\ss}ner, M.,
  Klein, R., Kolb, A.: State of the art on {3D} reconstruction with {RGB-D}
  cameras. {Computer Graphics Forum (CGF)}  \textbf{37}(2),  625--652 (2018)

\end{thebibliography}

\end{document}